%% file: acl_latex.tex
\documentclass{article}

\PassOptionsToPackage{numbers}{natbib}

\usepackage[preprint]{neurips_2026}

\usepackage{latexsym}
\usepackage{wrapfig}
\input{math_commands.tex}

\usepackage{times}
\usepackage{color}
\usepackage{bm}
\usepackage{tabularx}
\usepackage{tabularray}
\usepackage{dsfont}
\usepackage[T1]{fontenc}

\usepackage[utf8]{inputenc}

\usepackage[dvipsnames]{xcolor}
\definecolor{linkColor}{rgb}{0.18,0.39,0.62}
\usepackage[colorlinks=true,linkcolor=linkColor,citecolor=linkColor,filecolor=linkColor,urlcolor=linkColor]{hyperref}

\usepackage{microtype}

\usepackage{inconsolata}

\usepackage{graphicx}
\usepackage{microtype}

\usepackage{caption}
\usepackage{hyperref}       %
\usepackage{inconsolata}
\usepackage{algorithm}
\usepackage{wrapfig}
\usepackage{algorithmic}
\usepackage{colortbl}
\usepackage{placeins}

\usepackage{wrapfig}
\usepackage{xcolor}
\usepackage{inconsolata}
\usepackage{multirow}
\usepackage{amssymb}
\usepackage{amsmath}
\usepackage{color}
\usepackage{fontawesome5}
\usepackage{subfigure}
\usepackage{makecell}
\usepackage{booktabs}
\usepackage{enumitem}
\usepackage{multirow}

\usepackage{booktabs}
\usepackage{stfloats}
\usepackage{listings}
\usepackage{cite}
\usepackage{pifont}
\usepackage{tcolorbox}
\usepackage{soul}
\usepackage{enumitem}
\tcbuselibrary{skins, breakable, theorems}
\usepackage{wrapfig}
\usepackage{lscape}
\usepackage{url}
\usepackage{tikz}

\definecolor{mypink}{rgb}{.99,.91,.95}
\definecolor{myyellow}{rgb}{.99,.94,.82}
\newcommand{\ours}[0]{{Ctx2Skill}}

\newcommand{\skillR}[1]{\mathcal{S}^{\mathrm{R}}_{#1}}
\newcommand{\skillC}[1]{\mathcal{S}^{\mathrm{C}}_{#1}}
\newcommand{\probeH}{\mathcal{Q}^{\mathrm{h}}}
\newcommand{\probeE}{\mathcal{Q}^{\mathrm{e}}}

\NewDocumentCommand{\zhenhailong}
{ mO{} }{\textcolor{purple}{\textsuperscript{\textit{Zhenhailong}}\textsf{\textbf{\small[#1]}}}}

\input{config}

\title{From Context to Skills: Can Language Models \\ Learn from Context Skillfully?}

\author{
\textbf{Shuzheng Si\footnotemark[1]~~$^{\spadesuit\diamondsuit}$, Haozhe Zhao\footnotemark[1]~~$^{\clubsuit}$, Yu Lei\footnotemark[1]~~$^{\diamondsuit}$, Qingyi Wang\footnotemark[1]~~$^{\bigstar}$, Dingwei Chen$^{\heartsuit}$} \\
\textbf{Zhitong Wang$^{\spadesuit}$, Zhenhailong Wang$^{\clubsuit}$, Kangyang Luo$^{\spadesuit}$, Zheng Wang$^{\clubsuit}$} \\
\textbf{Gang Chen$^{\diamondsuit}$, Fanchao Qi$^{\diamondsuit}$, Minjia Zhang$^{\clubsuit}$,} and \textbf{Maosong Sun$^{\spadesuit}$} \\ 
$^{\spadesuit}$THU \quad $^\diamondsuit$DeepLang AI \quad $^{\clubsuit}$UIUC \quad $^{\bigstar}$FDU  \quad $^{\heartsuit}$CUHK \\
[0.2em]
{\ttfamily\small
\raisebox{-0em}{\faGithub}\ \href{https://github.com/S1s-Z/Ctx2Skill}{S1s-Z/Ctx2Skill}
}
\quad \small{$^{*}$~Equal Contribution.}
}

\begin{document}

\maketitle

\renewcommand{\thefootnote}{\fnsymbol{footnote}}
\renewcommand{\thefootnote}{\arabic{footnote}}
\urlstyle{same}
\definecolor{darkgreen}{RGB}{50,100,0}
\definecolor{darkred}{RGB}{200, 0, 0}
\definecolor{lightred}{RGB}{250, 200, 200}
\definecolor{lightblue}{RGB}{210, 220, 250}
\newcommand{\cmark}{\textcolor{darkgreen}{\ding{51}}} %
\newcommand{\xmark}{\textcolor{darkred}{\ding{55}}} %
\definecolor{tabcolor1}{RGB}{247,225, 237} 
\definecolor{tabcolor2}{RGB}{255, 250, 132} 
\definecolor{tabcolor3}{RGB}{204, 232, 207} 
\definecolor{tabcolor4}{RGB}{245, 222, 179} 
\definecolor{tabcolor5}{RGB}{210, 220, 250} 
\definecolor{tabcolor6}{RGB}{222, 222, 222} 

\vspace{-2.5mm}

\input{Files/0_Abstract}
\input{Files/1_Introduction}
\input{Files/2_Related_work}

\input{Files/3_Method}
\input{Files/4_Experiment}
\input{Files/5_Conclusion}

\bibliographystyle{acl_natbib}
\bibliography{anthology,custom}

\include{Files/6_Appendix}

\end{document}

%% file: math_commands.tex
\usepackage{amsmath,amsfonts,bm}

\def\eqref#1{equation~\ref{#1}}

\def\1{\bm{1}}

\DeclareMathAlphabet{\mathsfit}{\encodingdefault}{\sfdefault}{m}{sl}
\SetMathAlphabet{\mathsfit}{bold}{\encodingdefault}{\sfdefault}{bx}{n}



%% file: config.tex
\definecolor{Gray}{gray}{0.9}
\definecolor{mygreen}{rgb}{0.0, 0.5, 0.0}
\definecolor{myred}{rgb}{0.8, 0.25, 0.33}
\definecolor{myblue}{rgb}{0.19, 0.55, 0.91}
\definecolor{uclablue}{rgb}{0.15, 0.45, 0.68}
\definecolor{ucladblue}{rgb}{0.0, 0.33, 0.53}
\definecolor{ucladdblue}{rgb}{0.0, 0.23, 0.36}
\definecolor{uclagold}{rgb}{1.0, 0.82, 0.0}
\definecolor{ucladgold}{rgb}{1.0, 0.78, 0.17}
\definecolor{ucladdgold}{rgb}{1.0, 0.72, 0.11}
\definecolor{boxgreen}{rgb}{0.02, 0.66, 0.02}
\definecolor{boxred}{rgb}{0.66, 0.1, 0.1}
\definecolor{boxblue}{rgb}{0.01, 0.01, 0.73}

\usepackage{times}
\usepackage{latexsym}

\usepackage[utf8]{inputenc}
\usepackage[T1]{fontenc}

\usepackage{inconsolata}

\usepackage{algorithm}
\usepackage{algorithmic}

\usepackage{newfloat}
\usepackage{listings}
\floatstyle{ruled}
\newfloat{listing}{tb}{lst}{}
\floatname{listing}{Listing}

\usepackage{graphicx}
\usepackage{amsmath,amssymb,amsthm,mathabx,amsfonts}
\usepackage{bbm}
\usepackage{cleveref}
\usepackage{acronym}
\usepackage{enumitem}
\usepackage{balance}
\usepackage{xspace}
\usepackage{setspace}
\usepackage{url}
\usepackage{amsfonts}
\usepackage{multirow}
\usepackage{booktabs}
\usepackage{tablefootnote}
\usepackage{multicol,lipsum}
\usepackage{tikz}
\usepackage{pgfplots}
\usepackage{pgfplotstable}
\usepackage{arydshln}

\usepackage{CJKutf8}

\pgfplotsset{compat=1.18}
\makeatletter
\DeclareRobustCommand\onedot{\futurelet\@let@token\@onedot}
\def\@onedot{\ifx\@let@token.\else.\null\fi\xspace}

\makeatother

\makeatletter
\newcommand{\thickhline}{%
    \noalign {\ifnum 0=`}\fi \hrule height 1pt
    \futurelet \reserved@a \@xhline
}

\crefname{algorithm}{Alg.}{Algs.}
\Crefname{algocf}{Algorithm}{Algorithms}
\crefname{section}{Sec.}{Secs.}
\Crefname{section}{Section}{Sections}
\crefname{table}{Tab.}{Tabs.}
\Crefname{table}{Table}{Tables}
\crefname{figure}{Fig.}{Fig.}
\Crefname{figure}{Figure}{Figure}
\crefname{appendix}{Appendix}{Appendices}

\acrodef{nlp}[NLP]{natural language processing}
\acrodef{plm}[PLM]{Pre-trained Language Model}
\acrodef{sota}[SOTA]{state-of-the-art}
\acrodef{icl}[ICL]{In-Context Learning}
\acrodef{bbl}[BBL]{BIG-bench Lite}

\usepackage{colortbl}
\definecolor{gblue}{HTML}{4285F4}
\definecolor{gred}{HTML}{DB4437}
\definecolor{ggreen}{HTML}{0F9D58}

\definecolor{mygray}{gray}{.92}
\definecolor{emphypurple}{rgb}{0.302, 0.055, 0.659}

\definecolor{highlightgreen}{HTML}{009901}
\definecolor{highlightred}{HTML}{FD6864}


%% file: Files/0_Abstract.tex
\begin{abstract}
Many real-world tasks require language models (LMs) to reason over complex contexts that exceed their parametric knowledge.
This calls for \textit{context learning}, where LMs directly learn relevant knowledge from the given context.
An intuitive solution is inference-time skill augmentation: extracting the corresponding rules and procedures from \textit{context} into explicit, natural-language \textit{skills}. 
However, constructing such skills for context learning scenarios faces two fundamental challenges: the prohibitive cost of manual skill annotation for long, technically dense contexts, and the lack of external feedback for automated skill construction.
In this paper, we propose \textbf{\textit{\ours}}, a self-evolving framework that autonomously discovers, refines, and selects context-specific skills without human supervision or external feedback. 
At its core, a multi-agent self-play loop has a \textit{Challenger} that generates probing tasks and rubrics, a \textit{Reasoner} that attempts to solve them guided by an evolving skill set, and a neutral \textit{Judge} that provides binary feedback. 
Crucially, both the \textit{Challenger} and the \textit{Reasoner} evolve through accumulated natural-language skills: dedicated \textit{Proposer} and \textit{Generator} agents analyze failure cases and synthesize them into targeted skill updates for both sides, enabling automated skill discovery and refinement.
To prevent adversarial collapse caused by increasingly extreme task generation and over-specialized skill accumulation, we further introduce a \textit{Cross-time Replay} mechanism that identifies the skill set achieving the best balance across representative cases for the \textit{Reasoner} side, ensuring robust and generalizable skill evolution. 
The resulting skills can be plugged into any language model to obtain better context learning capability. 
Evaluated on four context learning tasks from CL-bench, \textit{\ours} consistently improves solving rates across backbone models, e.g., lifting GPT-4.1 from $11.1\%$ to $16.5\%$, GPT-5.1 from $21.1\%$ to $25.8\%$.
\end{abstract}

%% file: Files/1_Introduction.tex
\section{Introduction}
\label{section:introduction}

Current language models (LMs) have achieved impressive performance on problems whose relevant knowledge was present during large-scale pre-training~\citep{Claude4, gpt-5, team2026qwen3}, such as competition-level mathematical problems \citep{gao2025omnimath} and competitive programming challenges \citep{jimenez2024swebench}.
However, many real-world tasks require LMs to learn from complex contexts and leverage new knowledge rather than parametric knowledge to reason and solve them effectively \citep{Mei2025ASO, canoe, zhang2026agenticcontextengineeringevolving}.
Previous works refer to this capability as \textbf{\textit{context learning}} \citep{dou2026clbench}.
Effective context learning enables models to reason beyond their pre-trained knowledge and solve complex, domain-specific tasks by learning directly from rich contextual information, much as humans do.
For example, it allows LMs to rapidly make use of previously unseen product documentation to generate step-by-step operational procedures or troubleshoot issues.

Despite the central role of context learning in solving real-world tasks, it has been largely overlooked in current research \citep{hua2025contextengineering20context}.
A key obstacle is the diversity of real-world contexts: contextual knowledge can be conveyed through books, experimental data, and search results, among others, making context learning fundamentally difficult.
An intuitive paradigm is inference-time skill augmentation \citep{a-skills-2}, i.e., extracting rules and procedures from \textbf{\emph{context}} into natural-language \textbf{\emph{skills}} that encode reusable procedural knowledge for LMs. 
This approach has proven effective in many agent tasks \citep{wang2026webxskillskilllearningautonomous, zheng2025skillweaverwebagentsselfimprove, si2025goalplanjustwish}, e.g., coding agents \citep{ma2026scalingcodingagentsatomic}.
However, this paradigm faces two fundamental challenges in context learning scenarios.
Specifically, \textbf{\emph{(1) Prohibitive Cost for Manual Skill Annotation}}:
Existing agent skill libraries are predominantly constructed through human annotation \citep{li2026organizingorchestratingbenchmarkingagent, liang2026skillnetcreateevaluateconnect, wang2026skillorchestralearningrouteagents}. 
However, this paradigm breaks down in context learning scenarios, where contexts are long, technically dense, and domain-specific. 
Curating high-quality skills requires annotators to fully internalize complex, multi-section documents, a process that is both cognitively demanding and economically infeasible, rendering human-curated skill construction impractical at scale.
\textbf{\emph{(2) Lack of External Feedback for Automated Skill Construction}}: 
Unlike verifiable tasks such as coding or mathematical reasoning \citep{cobbe2021training, shao2024deepseekmath}, where skill quality can be assessed via execution feedback or ground-truth comparison, context learning tasks lack automatic feedback to evaluate whether extracted skills faithfully and completely capture context-specific knowledge.
Given only the context, there is no external feedback signal to tell whether a generated skill is useful or whether critical knowledge has been omitted.
This feedback-free nature renders automated skill construction pipelines inapplicable \citep{alzubi2026evoskill, yang2026autoskill, zhang2026memskilllearningevolvingmemory}.
\textit{Together, these challenges highlight the need for an automated framework that discovers skills from the context without relying on either human annotation or external feedback signals.}

\begin{figure}[t]
    \centering
    \includegraphics[width=\linewidth]{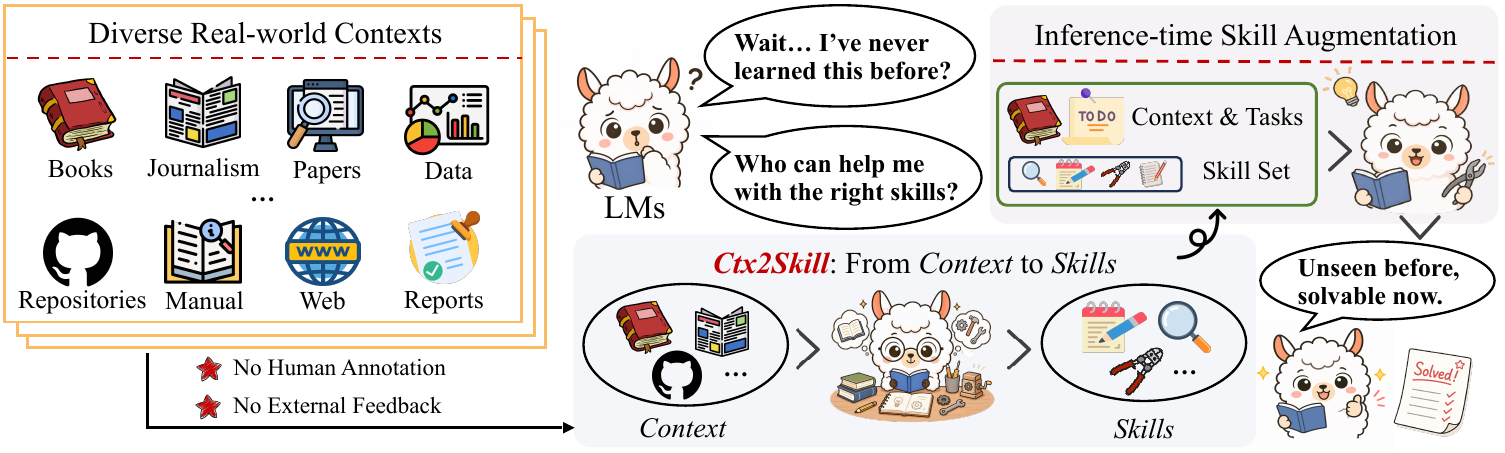}
    \vspace{-4.7mm}
    \caption{\textbf{The illustration of \textit{\ours{}}.} It's designed to extract rules and procedures from \textit{context} into natural-language \textit{skills} without human annotation and external feedback.}
    \label{figure:intro}
    \vspace{-4.5mm}
\end{figure}

In this paper, we propose \textbf{\textit{\ours}} shown in Figure \ref{figure:intro}, which is designed to autonomously discover, refine, and select skills directly from complex contexts via skill-optimized self-play, requiring neither human annotation nor external feedback.
At the core of \textit{\ours} is a multi-agent self-play loop \citep{digiovanni2021surveyselfplayreinforcementlearning, lu2026search, zhang2026piplaymultiagentselfplayprivileged, liu2026spiralselfplayzerosumgames} comprising two competing but co-evolving roles: a \textit{Challenger} agent, which generates a batch of tasks and associated rubrics \citep{gunjal2025rubricsrewardsreinforcementlearning} based on context and its skill set, aiming to probe deep contextual understanding, and a \textit{Reasoner} agent, which reads the context and attempts to solve these tasks guided by its current skill set.
The core idea of \textit{\ours{}} is that the two agents co-evolve by iteratively updating their respective skill sets rather than model parameters.
Specifically, a neutral \textit{Judge} agent evaluates the \textit{Reasoner}'s responses against each rubric, producing feedback on whether it passes.
Crucially, both the \textit{Challenger} and \textit{Reasoner} evolve through accumulated natural-language skills: failed cases are routed to dedicated \textit{Proposer} and \textit{Generator} agents on the \textit{Reasoner} side to diagnose missing contextual knowledge and update skills accordingly, while easily solved cases are routed to the \textit{Challenger} side to strengthen its task and rubric generation strategies, ensuring sustained adversarial pressure.
The updated skills are carried forward into the next iteration, allowing both sides to continuously co-evolve through failure-driven textual feedback without any parameter updates.
Furthermore, we identify a key risk in this multi-agent self-play loop, namely \textit{adversarial collapse} phenomenon: as iterations progress, the \textit{Challenger} may generate increasingly extreme tasks while the \textit{Reasoner}'s skills over-specialize to these pathological cases, resulting in redundant skill accumulation that degrades generalization.
To address this, we further propose a \textit{Cross-time Replay} mechanism that selects the skill set that best balances performance across representative cases, ensuring robust and generalizable skill evolution.
At inference time, the resulting context-specific \textit{Reasoner} skills can be plugged into any LM to enhance its context learning capability on arbitrary tasks over the same context, including those unseen during the self-play loop.

We evaluate our \textit{\ours{}} on four context learning tasks from CL-Bench~\citep{dou2026clbench}, including {Domain Knowledge Reasoning}, {Rule System Application}, {Procedural Task Execution}, and {Empirical Discovery \& Simulation}.
\textit{\ours{}} consistently improves solving rate across all four categories and multiple backbone models: it raises GPT-4.1 \citep{gpt-4-tr} from $11.1\%$ to $16.5\%$, GPT-5.1 \citep{gpt-5} from $21.1\%$ to $25.8\%$, and GPT-5.2 \citep{gpt-5} from $18.2\%$ to $21.4\%$, demonstrating that autonomously discovered context-specific skills provide substantial and generalizable gains.

%% file: Files/2_Related_work.tex
\section{Related Work}
\label{section:related_work}

\textbf{Context Learning.}
Current language models excel at tasks whose relevant knowledge was present during pre-training \citep{deepseekai2025deepseekv3technicalreport, gpt-5, si2026faithlensdetectingexplainingfaithfulness, si-etal-2025-aligning-acl}.
However, many real-world tasks require models to learn from complex contexts and leverage new knowledge to reason effectively, a capability termed \emph{context learning}~\citep{dou2026clbench}.
This context learning capability is essential in practice: a physician consults newly released clinical guidelines to adjust treatment, an engineer reads documentation to execute procedures, and a researcher induces patterns from experimental data to form hypotheses.
However, prior work finds that LMs have yet to master this capability \citep{dou2026clbench}.
The core difficulty is that contextual knowledge spans diverse formats, from books to experimental datasets, and the relevant rules and procedures are often embedded implicitly, requiring deep comprehension rather than surface-level retrieval.
\textit{\ours} directly targets this challenge by autonomously discovering reusable, context-specific skills from the contexts, equipping any LM to better learn from and reason over complex contexts.

\textbf{Skills for LMs.}
Skills are natural-language modules that encode reusable procedural knowledge to augment LMs at inference time \citep{a-skills-1, a-skills-2}, and skill augmentation has been widely validated across agent tasks \citep{si2023spokenwoz}, e.g., coding \citep{ma2026scalingcodingagentsatomic} and web navigation \citep{wang2026webxskillskilllearningautonomous}.
Early skill libraries are predominantly constructed through human annotation \citep{li2026organizingorchestratingbenchmarkingagent, li2026skillsbenchbenchmarkingagentskills, liang2026skillnetcreateevaluateconnect, wang2026skillorchestralearningrouteagents}, which is effective but does not scale, especially for context learning scenarios where contexts are long, technically dense, and domain-specific.
Recent work has shifted toward automated skill construction.
AutoSkill \citep{{yang2026autoskill}} extracts reusable behaviors from interaction traces as lifelong versioned artifacts, and AutoRefine \citep{qiu2026autorefine} converts agent trajectories into reusable expertise.
CoEvoSkills \citep{zhang2026coevoskillsselfevolvingagentskills} generates multi-file skill packages and refines them through generator--verifier co-evolution, while EvoSkill \citep{alzubi2026evoskill} performs failure-driven refinement into structured skill folders.
SkillX \citep{wang2026skillx} distills the agent trajectories into a hierarchical skill knowledge base and iteratively refines them through execution feedback.
However, these methods rely on external feedback signals \citep{ma2026skillclawletskillsevolve}, such as execution feedback, ground-truth comparison, or task-completion rewards, to evaluate and improve skill quality, which are not available in context learning scenarios where there is no automatic feedback.
A separate line of work tries to internalize skills into model parameters.
SKILL0 \citep{lu2026skill0} uses in-context reinforcement learning (RL) to absorb skills, and SkillRL \citep{xia2026skillrlevolvingagentsrecursive} builds a hierarchical skill bank via RL-guided distillation from teacher trajectories.
These approaches require parameter access, making them inapplicable to closed-source models, and sacrificing the interpretability of natural-language skill documents.
In this paper, we propose \textit{\ours}, a self-evolving framework designed to directly discover and evolve skills from complex contexts alone, which requires no human annotation, no external feedback, and no parameter updates.

%% file: Files/3_Method.tex
\section{Methodology}
\label{section:method}

\subsection{Overview}
\label{sec:overview}

\begin{figure}[t]
    \centering
    \includegraphics[width=\linewidth]{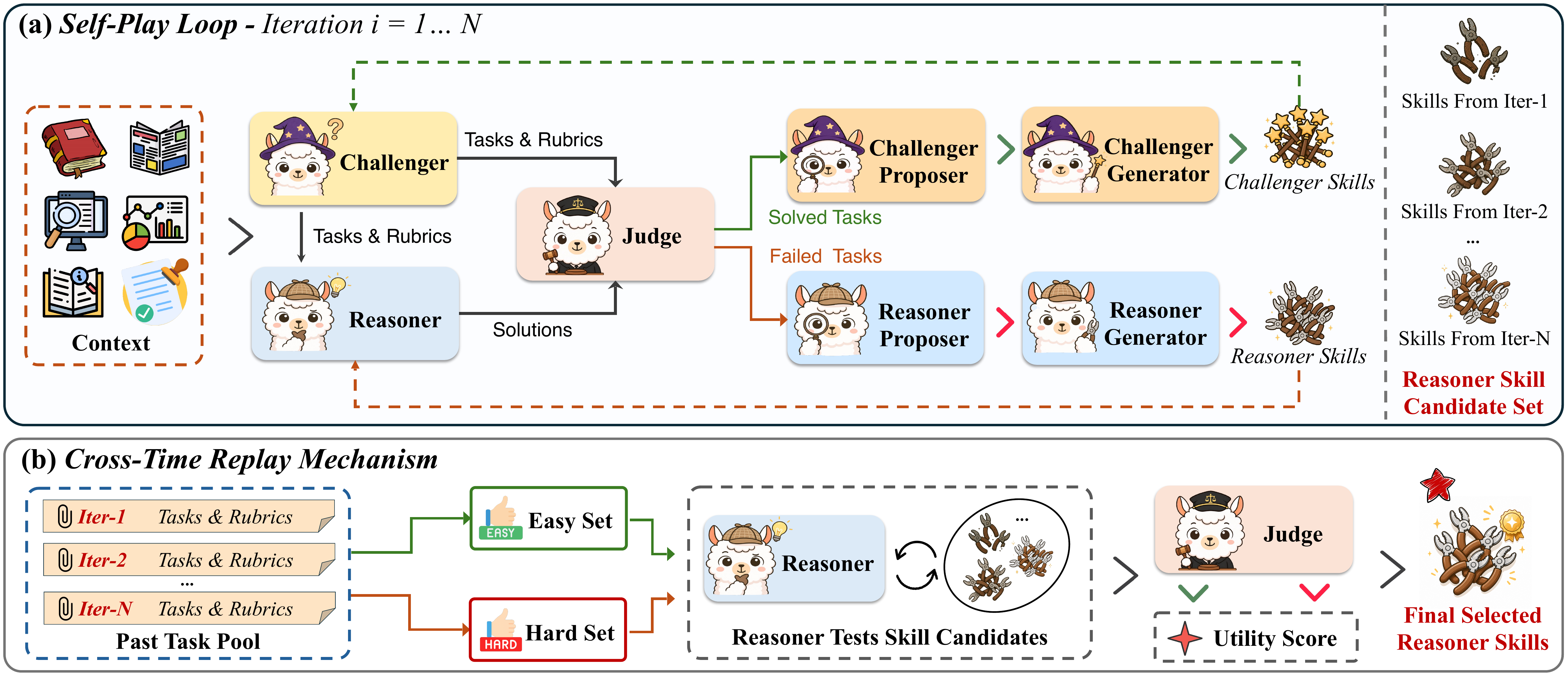}
    \vspace{-4.7mm}
    \caption{\textbf{The overview of \textit{\ours{}}}. (a) In the self-play loop, the \textit{Challenger} generates tasks and rubrics, the \textit{Reasoner} tries to solve them, and the \textit{Judge} routes outcomes to update \textit{Challenger} and \textit{Reasoner} skills.
    (b) Cross-Time Replay Mechanism re-evaluates historical \textit{Reasoner} skill candidates on representative tasks, selecting the most balanced skill set for unseen downstream tasks.}
    \label{figure:method}
    \vspace{-4.5mm}
\end{figure}

Context learning tasks \citep{dou2026clbench} require LMs to leverage contextual knowledge from long, technically dense contexts to reason and solve tasks effectively.
An intuitive paradigm for context learning tasks is inference-time skill augmentation, i.e., extracting the corresponding rules and procedures from \textbf{\emph{context}} into natural-language \textbf{\emph{skills}}.
Turning this idea into a working framework, however, faces two difficulties: (1) prohibitive cost for manual skill annotation and (2) lack of external feedback for automated skill construction.
In this paper, we introduce \textbf{\textit{\ours}} shown in Figure \ref{figure:method}, a self-evolving framework that autonomously discovers, refines, and selects context-specific skills without human supervision or external feedback. 
At its core, a \emph{skill-optimized self-play} loop (\S~\ref{sec:selfplay}) co-evolves two competing sides: a \textit{Challenger} probes the context with tasks and rubrics guided by its own skill set, a \textit{Reasoner} solves them under its current skill set, and a neutral \textit{Judge} returns binary verdicts that route each outcome to a side-specific \textit{Proposer}--\textit{Generator} pair, which diagnoses the exposed weakness and edits that side's skills accordingly, so that both sides co-evolve without human annotation or external feedback.
Furthermore, a key risk in this self-play framework is \emph{adversarial collapse}: the \textit{Challenger} may generate increasingly extreme tasks while the \textit{Reasoner}'s skills over-specialize to these pathological cases, resulting in redundant skill accumulation that degrades generalization.
To address this, we further propose a \textit{Cross-Time Replay} mechanism (\S~\ref{sec:crosstime}) that selects the skill set that best balances performance across representative cases, ensuring robust and generalizable skill evolution.
In this way, \textit{\ours} autonomously constructs context-specific skills that can be plugged into any LM at inference time to enhance context learning capability without parameter updates.

\subsection{Problem Formulation}
\label{sec:setting}
A context learning task \citep{dou2026clbench} consists of a given context $C$ whose content may lie outside the LM's pretraining corpus, a set of tasks $\mathcal{T}=\{t_j\}$ whose correct answers depend on $C$, and a set of binary rubrics $\mathcal{R}_j=\{r_{j,k}\}$.
Given an answer $a_j$ produced by an LM $\pi$, the task is considered solved only when every rubric passes, we define a solving indicator $y_j(\pi; C)$ for task $t_j$ as:
\begin{equation}
\label{eq:task_score}
y_j(\pi;C) = \prod_{k} \mathbb{I}\bigl[r_{j,k}(a_j)=\mathrm{pass}\bigr],
\qquad a_j \sim \pi(\cdot\mid C, t_j).
\end{equation}
Our goal is to enable the LM $\pi$ to solve tasks over a previously unseen context $C$ without any parameter update or external feedback.
To this end, we introduce a natural-language \emph{skill set} $\mathcal{S}$, a short Markdown file pre-pended to the system prompt of the LM at inference time:
\begin{equation}
\label{eq:skill_extended}
a_j \sim \pi(\cdot\mid \mathcal{S}, C, t_j).
\end{equation}
During the skills construction of our proposed \textit{\ours{}}, $\mathcal{S}$ is instantiated as two separate skill sets: $\skillR{}$ for the \textit{Reasoner} agent and $\skillC{}$ for the \textit{Challenger} agent.
At inference time, only the \textit{Reasoner}'s skill set $\skillR{}$ is deployed with an LM for context-specific tasks evaluation.

\subsection{\textit{Ctx2Skill} Framework}
\label{sec:selfplay}
Building a skill set for an unseen context $C$ faces two obstacles: human annotation is costly to read $C$ and write down its contextual knowledge, and no external feedback signal can tell us whether a proposed skill is faithful and useful to $C$. 
To address both, \textit{Ctx2Skill} runs a multi-agent self-play loop that autonomously discovers and refines context-specific skills from context $C$ alone. 

The core idea of \textit{Ctx2Skill} is to co-evolve two skill sets, skill set $\skillR{}$ for the \textit{Reasoner} and skill set $\skillC{}$ for the \textit{Challenger}, through $N$ iterations of failure-driven textual edits, so that the \textit{Reasoner} gradually accumulates the contextual knowledge of the context $C$ while the \textit{Challenger} keeps probing aspects not yet mastered. 
Five frozen-LM agent roles carry out this self-play loop. 
At iteration $i$, given the context $C$ and skill set $\skillC{i-1}$, the \emph{Challenger} produces a batch of tasks with corresponding rubrics; the \emph{Reasoner}, guided by context $C$ and its skills $\skillR{i-1}$, attempts each task; the neutral \emph{Judge} returns per-task binary verdicts and partitions the batch into failed and solved tasks. 
Each group is then routed to its corresponding side through a \emph{Proposer}--\emph{Generator} pair: failed cases flow to the \emph{Reasoner} side, where the \textit{Proposer} diagnoses which contextual knowledge is missing and the \emph{Generator} updates $\skillR{}$ accordingly; solved cases flow to the \emph{Challenger} side, where the corresponding pair tightens $\skillC{}$ so that the next iteration sustains adversarial pressure.
The updated $\skillR{i}$ and $\skillC{i}$ are carried into iteration $i{+}1$, and no prompt is ever conditioned on the opposing side's skill set, keeping the two sides strictly adversarial.
We now define the five agent roles as follows at the iteration $i$:

\begin{itemize}[leftmargin=1.4em,itemsep=3pt,topsep=3pt]
    \item \emph{\textbf{Challenger}}: Given the context $C$ and its current skill set $\skillC{i-1}$, the \textit{Challenger} produces a batch of $M$ tasks $\{t_m\}_{m=1}^M$, each equipped with corresponding rubrics $\mathcal{R}_m$. The rubrics are designed so that a correct answer requires inducing the rules of $C$ rather than paraphrasing surface spans.
    By conditioning on its skills $\skillC{i-1}$, the \textit{Challenger} adapts its probing strategy across iterations, sustaining adversarial pressure as the \textit{Reasoner} improves.

    \item \emph{\textbf{Reasoner}}: Given $C$, a task $t_m$, and its current skill set $\skillR{i-1}$, the \textit{Reasoner} produces an answer $a_m$.
    While the raw context $C$ contains all necessary knowledge, it is often long and technically dense; the skill set $\skillR{i-1}$ distills the most relevant rules and procedures from $C$ into a concise, structured form, allowing the \textit{Reasoner} to directly apply contextual knowledge that would otherwise need to be extracted from scratch at every task.

    \item \emph{\textbf{Judge}}: Given a task and rubrics generated from the \textit{Challenger}, and the generated answer from the \textit{Reasoner}, the \textit{Judge} returns a per-rubric binary verdict $z_{m,k} = \mathbb{I}[r_{m,k}(a_m) = \mathrm{pass}]$ and a solving indicator $y_m = \prod_k z_{m,k}$. The \textit{Judge} is a strict evaluator: the task is considered solved if and only if the answer satisfies all associated rubrics.
    
    \item \emph{\textbf{Proposer} (one per side)}:   The \textit{Judge}'s binary verdicts identify \emph{which} tasks failed or solved, but not \emph{why}. 
    The \textit{Proposer} therefore collects all routed cases of its side, where each case comprises the task and rubrics from the \textit{Challenger}, and the \textit{Reasoner}'s answer, and jointly analyzes them against the current skill set. 
    Rather than diagnosing each case in isolation, the \textit{Proposer} synthesizes common failure or success patterns across cases into a high-level diagnosis specifying an action (add or merge), the target skill name, a description, and a justification.
    The \emph{\textbf{Reasoner Proposer}} is designed to take the failed cases $\mathcal{F}_i$ together with $\skillR{i-1}$, and diagnose which contextual knowledge is missing or misrepresented by examining both the incorrect answers and the rubrics they violated.
    On the other hand, the \emph{\textbf{Challenger Proposer}} takes the easily solved cases $\mathcal{P}_i$ together with $\skillC{i-1}$, and identifies gaps in the current task and rubric generation strategy by inspecting how the \textit{Reasoner} solved them with its existing skills.
    
    \item \emph{\textbf{Generator} (one per side)}:
    The \textit{Proposer}'s output is a high-level diagnosis describing what should change and why, but not the concrete skill content itself. 
    The \textit{Generator} materializes this diagnosis into an actual skill set. 
    Given the diagnosis and the current skill set, it returns a complete replacement skill set that adds or merges entries while preserving every unrelated entry.
    The \textbf{\textit{Reasoner Generator}} incorporates the diagnosed missing contextual knowledge into $\skillR{i-1}$ to produce the new skill set $\skillR{i}$.
    The \textbf{\textit{Challenger Generator}} tightens $\skillC{i-1}$ so that the next iteration probes the \textit{Reasoner}'s remaining weaknesses, producing skills $\skillC{i}$.
\end{itemize}

To summarize, at iteration $i$, the \textit{Challenger} first produces a batch of tasks and rubrics $\{(t_m, \mathcal{R}_m)\}_{m=1}^M$ based on the context $C$; the \textit{Reasoner} attempts to answer each $t_m$; the \textit{Judge} evaluates every rubric and partitions the batch into failed cases $\mathcal{F}_i = \{m : y_m = 0\}$ and solved cases $\mathcal{P}_i = \{m : y_m = 1\}$ based on the solving indicator $y_m$. Each group is then routed to its corresponding side:
\begin{itemize}[leftmargin=1.4em,itemsep=1pt,topsep=2pt]
    \item The \textit{Reasoner Proposer} jointly analyzes all failed cases $\mathcal{F}_i$ against $\skillR{i-1}$ and synthesizes a high-level diagnosis; the \textit{Reasoner Generator} then materializes it into $\skillR{i}$.
    \item The \textit{Challenger Proposer} jointly analyzes all solved cases $\mathcal{P}_i$ against $\skillC{i-1}$ and synthesizes a high-level diagnosis; the \textit{Challenger Generator} then materializes it into $\skillC{i}$.
\end{itemize}
The updated skill sets $\skillR{i}$ and $\skillC{i}$ are carried forward into iteration $i{+}1$. No prompt is ever conditioned on the opposing side's skill set, keeping the two sides strictly adversarial.

\subsection{\textit{Cross-Time Replay} Mechanism}
\label{sec:crosstime}
The framework of \textit{\ours{}} detailed in \S~\ref{sec:selfplay} intentionally strengthens the \textit{Challenger} across iterations through updates to $\skillC{}$, sustaining adversarial pressure as the \textit{Reasoner} improves.
However, this design introduces an inherent tension that we term \emph{adversarial collapse}.
As iterations progress, the \textit{Challenger} generates increasingly extreme tasks that concentrate on the \textit{Reasoner}'s residual weaknesses, gradually deviating from the representative knowledge of the context $C$.
Since the \textit{Reasoner}'s skill updates are failure-driven, its skills over-specialize to these pathological cases, resulting in redundant skill accumulation that degrades generalization.
Moreover, this collapse is undetectable within the loop: each iteration's \textit{Judge} only evaluates the \textit{Challenger}'s newly generated tasks, providing no signal on whether contextual knowledge mastered in earlier iterations has been corrupted by subsequent edits.
Returning the last iteration \textit{Reasoner}'s skills $\skillR{N}$ unconditionally is therefore unreliable; we instead select the most generalizable skill set among candidates $\{\skillR{1}, \dots, \skillR{N}\}$.

To address this, we propose a \emph{cross-time replay} mechanism that selects the most generalizable skill set from the candidates $\{\skillR{1}, \dots, \skillR{N}\}$ using representative probes collected during self-play itself.
Two small probe sets are curated incrementally without any external supervision: at each iteration $i$, the failed task with the lowest rubric pass rate is added to the hard probe set $\mathcal{Q}^{\mathrm{h}}$, and the solved task with the fewest rubrics is added to the easy probe set $\mathcal{Q}^{\mathrm{e}}$, capturing the hardest failure and the easiest success observed at each iteration.
Once the self-play loop terminates, the \textit{Reasoner} $\pi^\mathrm{R}$ re-answers every task in both probe sets under each candidate skill set $\skillR{i}$, and the \textit{Judge} re-evaluates every rubric, yielding the solving indicator $y_q$ (i.e., whether all rubrics pass) for each probe task $q$. 
We define the Laplace-smoothed \citep{field1988laplacian} solving rates $\rho^{\mathrm{h}}(i)$ and $\rho^{\mathrm{e}}(i)$ as the proportion of tasks solved in the hard and easy probe sets under the context $C$ and skills $\skillR{i}$, respectively:
\begin{equation}
\label{eq:accs}
\rho^h(i) = \frac{\sum_{q \in \probeH} y_q (\pi^\mathrm{R};C, \skillR{i}) + 1}{|\probeH| + 1},
\qquad
\rho^e(i) = \frac{\sum_{q \in \probeE} y_q (\pi^\mathrm{R};C, \skillR{i}) + 1}{|\probeE| + 1}.
\end{equation}

The selected skill set is the one that maximizes the product of these two solving rates:
\begin{equation}
\label{eq:select}
\skillR{\star} = \skillR{i^{\star}},
\qquad
i^{\star} = \arg\max_{i} \left(\rho^\mathrm{h}(i)\cdot\rho^\mathrm{e}(i)\right).
\end{equation}
The multiplicative form is essential: a skill set that resolves late, idiosyncratic failures at the cost of regressing on easier tasks incurs a penalty on $\rho^\mathrm{e}(i)$ and is rejected; conversely, one that solves every easy probe but fails every hard one is rejected through $\rho^\mathrm{h}(i)$.
Meanwhile, Laplace smoothing keeps the product well-defined when a probe set is empty for a given iteration

At inference time, the final skill set $\skillR{\star}$ is prepended to the \textit{Reasoner}'s system prompt for any unseen task $t_{u}$ over context $C$, and the \textit{Reasoner} answers conditioned on $(\skillR{\star}, C, t_{u})$.
Since $\skillR{\star}$ encodes reusable contextual knowledge of $C$ rather than task-specific solutions, it generalizes to arbitrary tasks over the same context.
Moreover, $\skillR{\star}$ is produced once per context and reused across all $|\mathcal{T}|$ unseen tasks, so the \textit{\ours{}} cost amortizes over $|\mathcal{T}|$ with no additional cost per deployed query.

%% file: Files/4_Experiment.tex
\section{Experiments}
\label{sec:experiment}
In this section, we conduct experiments and analyses to show the advantages of our \textit{\ours{}}.

\subsection{Settings}
\label{sec:setup}

\noindent\textbf{Evaluation.}
Context learning is a newly emerging research topic, aiming to require LMs to learn from unseen context and leverage new knowledge beyond what was acquired during pre-training to reason and solve tasks \citep{dou2026clbench}.
This goes far beyond long-context tasks \citep{bai2024longbench_acl, si-etal-2025-gateau} that primarily test retrieval or reading comprehension, and in-context learning tasks \citep{icl-survey}, where models learn simple task patterns via instructions and demonstrations. 
We select the CL-bench~\citep{dou2026clbench} for our evaluation, which comprises $500$ complex contexts, $1{,}899$ tasks, and $31{,}607$ verification
rubrics, all crafted by experienced domain experts.
Each task is designed such that the new content required to resolve it is contained within the corresponding context.
These data are distributed across four categories: \emph{Domain Knowledge Reasoning}, \emph{Rule System Application}, \emph{Procedural Task Execution}, and \emph{Empirical Discovery \& Simulation}.
Scoring is strictly all-or-nothing: a task is considered solved only when every rubric passes.
We report both per-category and overall solving rates.
Rubric-level grading follows the original benchmark protocol, which employs GPT-5.1 as the LLM-as-a-judge verifier~\citep{zheng2023mtbench}; cross-verifier and human agreement with this verifier both exceed $90\%$~\citep{dou2026clbench}.

\input{Tabs/main_result}

\noindent\textbf{Baselines.}
We first evaluate a broad set of frontier LMs, including GPT-4.1 \citep{gpt-4-tr}, GPT-5.1 \citep{gpt-5}, GPT-5.2 \citep{gpt-5-2}, Claude Opus 4.5 \citep{Claude4.5}, Gemini 3 Pro \citep{gemini-3}, Kimi K2.5 \citep{kimiteam2026kimik25visualagentic}, and DeepSeek V3.2 \citep{deepseekai2025deepseekv3technicalreport}.
For models that support thinking mode, we use the default reasoning effort.
For skill-based methods, we 
\begin{wraptable}[16]{r}{5.6cm}
    \centering
    \caption{\textbf{The quality evaluation of generated skills.}
    We use GPT-4.1 as a judge to assess skills across five dimensions: conciseness, faithfulness, clarity, effectiveness, and reusability.
    The best result in each dimension is marked in \textbf{bold}.}
    \label{tb:quality}
    \vspace{-1.5mm}
    \input{Tabs/llm-as-a-judge}
\end{wraptable}
compare \textit{\ours{}} with two baselines: \textit{(1) Prompting} directly prompts the LM to read the context $C$ and generate a skill set in a single pass; the resulting skills are then pre-pended to LMs during inference.
\textit{(2) AutoSkill4Doc} is a variant of AutoSkill \citep{yang2026autoskill} designed for document-level contexts: it partitions the context $C$ into different windows, identifies the independent skills within each window, and recombines skills across windows to produce a skill set.
Both baselines use the same backbone for skill generation and inference as \textit{\ours{}}: for example, all GPT-4.1-based methods use GPT-4.1 for both skill construction and evaluation, ensuring a fair comparison.

\noindent\textbf{Implementation.}
We run $N{=}5$ self-play iterations per context with $M{=}5$ tasks per iteration.
The \textit{Challenger}, \textit{Reasoner}, \textit{Proposer}, and \textit{Generator} all use the same backbone as the corresponding series (e.g., GPT-4.1 for all four agents in the GPT-4.1-based methods), while the \textit{Judge} uses GPT-5.1, consistent with the CL-bench evaluation protocol~\citep{dou2026clbench}.
Due to API budget constraints, we do not explore larger $N$ or $M$; nonetheless, the most effective skill set is typically selected from early iterations (\S~\ref{sec:analysis}), and $M{=}5$ tasks per iteration provide enough cases for the \textit{Proposer} to synthesize meaningful diagnostic patterns.

\subsection{Results}
\label{sec:main}
Table~\ref{tab:cl_bench_performance} presents the main results on CL-bench.
Context learning remains challenging for current frontier LMs: even the best-performing model, GPT-5.1, achieves only a $21.1\%$ overall solving rate.
\textit{\ours{}} consistently improves solving rates across all three backbones, raising GPT-4.1 from $11.1\%$ to $16.5\%$ ($+5.4\%$), GPT-5.1 from $21.1\%$ to $25.8\%$ ($+4.7\%$), and GPT-5.2 from $18.2\%$ to $21.4\%$ ($+3.2\%$), outperforming both \textit{Prompting} and \textit{AutoSkill4Doc} by a large margin across all four categories.
The gains are especially pronounced on Procedural Task Execution and Empirical Discovery \& Simulation, which demand deeper procedural and inductive reasoning over the context.
In contrast, the two baselines provide only modest improvements and occasionally degrade individual categories (e.g., \textit{Prompting} decreases Rule System Application on GPT-4.1 by $2.5\%$), suggesting that single-pass skill extraction is insufficient for complex contextual knowledge.
Notably, GPT-4.1 with \textit{\ours{}} skills ($16.5\%$) surpasses stronger frontier models without skills, e.g., Gemini 3 Pro ($15.8\%$), demonstrating that context-specific skills can bridge substantial capability gaps.
Beyond solving rate, we assess the intrinsic quality of generated skills using GPT-4.1 as a judge across five dimensions (Table~\ref{tb:quality}).
\textit{\ours{}} achieves the highest average score across all three backbones, outperforming \textit{AutoSkill4Doc} by $+3.6$, $+2.1$, and $+2.3$ on GPT-4.1, GPT-5.1, and GPT-5.2, with the most notable improvements in faithfulness and clarity.
These results indicate that the iterative self-play loop produces skills that not only improve downstream reasoning but also present contextual knowledge in a well-structured, human-readable form amenable to inspection, editing, and reuse.

\input{Tabs/aba}

\subsection{Analysis}
\label{sec:analysis}

\begin{figure}[t]
    \centering
    \includegraphics[width=\linewidth]{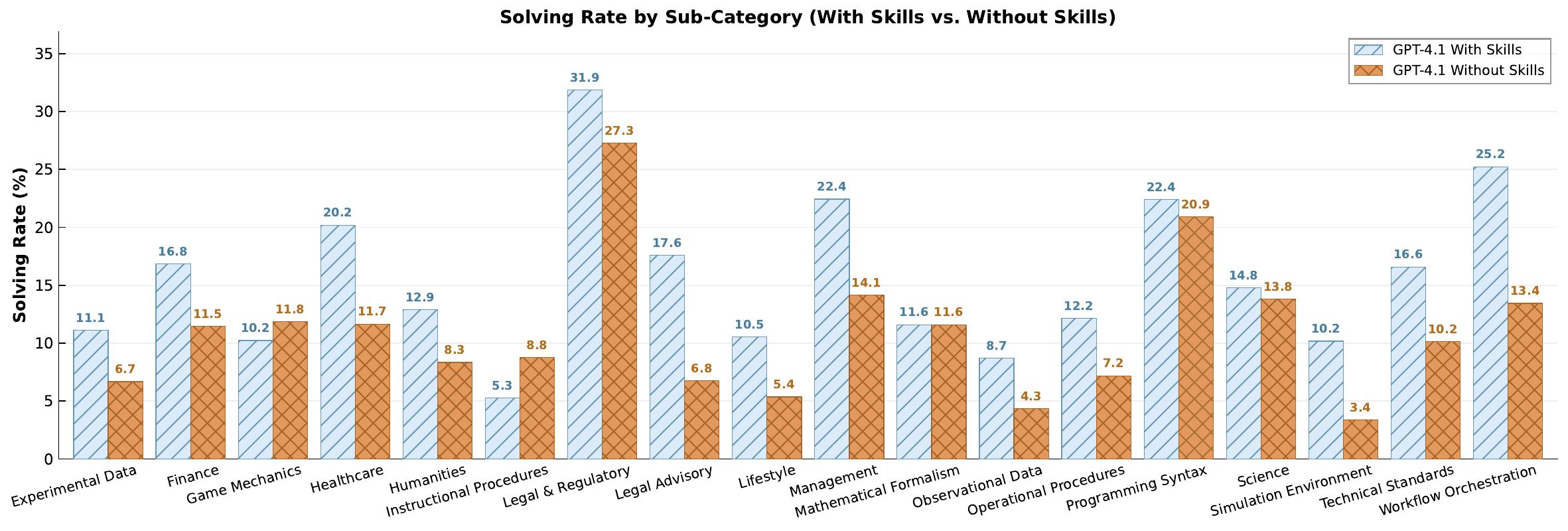}
    \vspace{-4.7mm}
    \caption{\textbf{Per sub-category solving rate on CL-bench.} \textit{\ours{}} improves solving rates on the vast majority of sub-categories compared with the base model without skills.}
    \label{figure:sub_cate}
    \vspace{-4.5mm}
\end{figure}

\textbf{Ablation Study.}
We ablate each component of \textit{\ours{}} on both the GPT-4.1 and GPT-5.1 (Table~\ref{tab:cl_bench_methods}, Ablation Study block).
Removing \textit{Challenger} skills evolving causes the largest drop, confirming that sustained adversarial pressure is essential for the \textit{Reasoner} to discover contextual knowledge progressively.
The \textit{Cross-Time Replay} mechanism is the second most impactful: without it, the last-iteration skills suffer from adversarial collapse.
Within \textit{Cross-Time Replay}, the hard probe set contributes more than the easy probe set, and removing Laplace smoothing also degrades performance.
Merging the \textit{Proposer} and \textit{Generator} into a single agent yields a modest but consistent decline, supporting the decoupling of diagnosis from skill materialization.
We also examine the impact across all sub-categories of CL-bench on GPT-4.1 (Figure~\ref{figure:sub_cate}).
\textit{\ours{}} improves
solving rates on the vast majority of sub-categories, with especially large gains on workflow orchestration ($+11.8\%$).

\textbf{Variant Designs Testing.}
We examine three alternative designs on GPT-4.1 (Table~\ref{tab:cl_bench_methods}, Variant Designs Testing block).
\textit{Loser-Only Skill Update} updates only the losing side each iteration; the $0.5\%$ drop 
\begin{wrapfigure}[16]{r}{6.5cm}
    \centering
    \vspace{-1.3pt}
    \includegraphics[width=\linewidth]{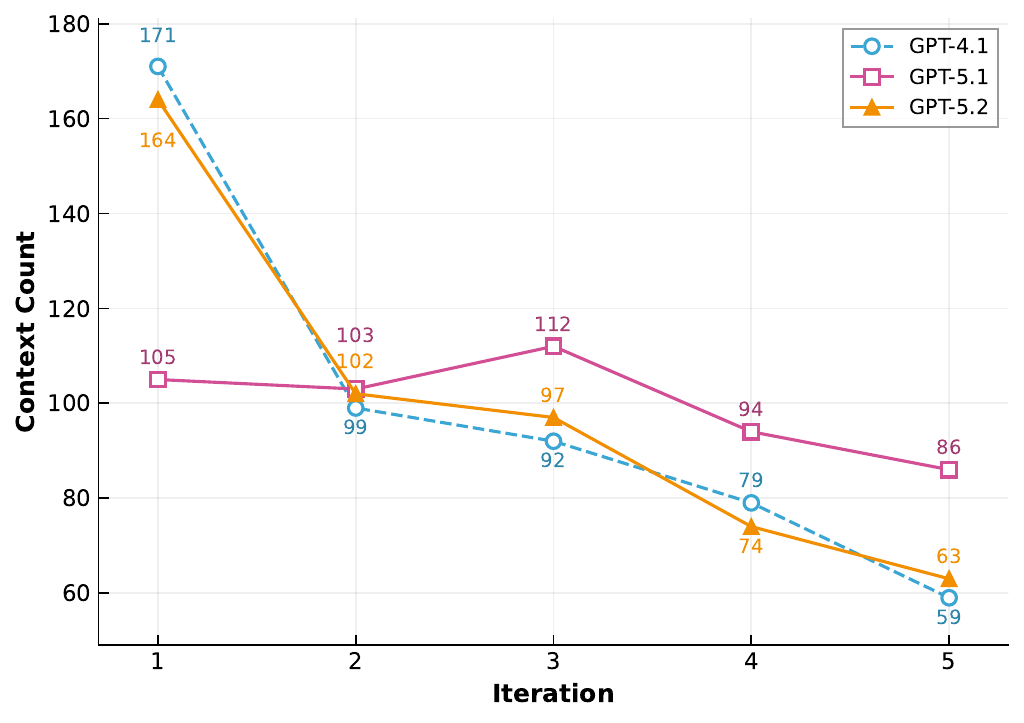}
    \vspace{-5.7mm}
    \captionof{figure}{\textbf{Distribution of selected iterations by the \textit{Cross-Time Replay}.} We report the number of contexts whose final skill set is selected from each iteration across all contexts.}
    \label{figure:skill_distribution}
\end{wrapfigure}
indicates that updating both sides yields more effective co-evolution.
\textit{Joint Outcome Skill Update} feeds both failed and solved cases into both sides simultaneously, allowing each side to learn from both positive and negative examples; the larger $1.0\%$ degradation suggests that mixing outcomes dilutes the diagnostic signal.
We also test \textit{Additive Scoring}, which replaces $\rho^\mathrm{h}(i) \cdot \rho^\mathrm{e}(i)$ with $\rho^\mathrm{h}(i) + \rho^\mathrm{e}(i)$ in Eq.~(\ref{eq:select}); the $0.6\%$ drop confirms that the multiplicative form better penalizes skill sets that sacrifice easy-probe performance for hard-probe gains.

\textbf{Effect of \textit{Cross-Time Replay} Mechanism.}
We compare \textit{Cross-Time Replay} against using a fixed iteration's skills (Table~\ref{tab:cl_bench_methods}, Effect of Cross-Time Replay block).
On GPT-4.1, fixed-iteration performance monotonically decreases from Iter-1 ($15.9\%$) to Iter-5 ($14.7\%$), confirming that later iterations suffer from adversarial collapse phenomenon.
\textit{Cross-Time Replay} ($16.5\%$) outperforms every fixed iteration, including the best one (Iter-1), by $+0.6\%$, by adaptively selecting the most balanced skill set per context rather than committing to a single iteration across all contexts.
Figure~\ref{figure:skill_distribution} corroborates this pattern: early iterations are selected most frequently across all three backbones, while a non-trivial proportion of later iterations indicates that certain contexts with more complex knowledge structures do benefit from additional self-play rounds.

\textbf{Skill Transferability Testing.}
We investigate whether skills generated by one backbone can benefit a different one (Table~\ref{tab:cl_bench_methods}, Skill Transferability Testing block).
GPT-5.1-generated skills applied to GPT-4.1 yield $16.1\%$, nearly matching GPT-4.1's own skills; conversely, GPT-4.1-generated skills applied to GPT-5.1 yield $23.1\%$, a $+2.0\%$ gain but notably below the $+4.6\%$ from GPT-5.1's own skills.
This asymmetry shows that stronger models produce skills that transfer well to weaker models, whereas weaker models lack the capacity to discover knowledge that stronger models can exploit.

\noindent
\textbf{Case Study \& Discussion.}
We provide a case study in Appendix \ref{appendix:case}.
We also discuss potential concerns about \textit{\ours{}}, e.g., reward hacking, in Appendix~\ref{appendix:discussion}.

%% file: Tabs/main_result.tex
\begin{table*}[t]
\caption{
\textbf{The main results of \textit{\ours{}}.}
We report the task-solving rate on the CL-bench across four task categories.
The best results within the same backbone are marked in \textbf{bold}.
{\small \faToggleOn} represents methods using inference time augmentation, and {\small \faToggleOff} shows inference results without skills.
\textcolor[rgb]{0.7,0,0}{Red} and \textcolor[rgb]{0,0.7,0}{Green} show the performance gains and declines, respectively, relative to the models without skills.}
\vspace{-1.5mm}
\centering
\resizebox{0.96\linewidth}{!}{
\begin{tabular}{lccccc}
\toprule
\textbf{Model} & 
\multicolumn{1}{c}{\textbf{Overall (\%)}} &
\multicolumn{1}{c}{\makecell{\textbf{Domain} \\ \textbf{Knowledge} \\ \textbf{Reasoning (\%)}}} & 
\multicolumn{1}{c}{\makecell{\textbf{Rule System} \\ \textbf{Application (\%)}}} &
\multicolumn{1}{c}{\makecell{\textbf{Procedural Task} \\ \textbf{Execution (\%)}}} & 
\multicolumn{1}{c}{\makecell{\textbf{Empirical} \\ \textbf{Discovery \&} \\ \textbf{Simulation (\%)}}} \\
\midrule

\multicolumn{6}{c}{\textbf{
Frontier LMs}} \\

\midrule

{\small \faToggleOff} GPT-5.1 &  {21.1} &  {22.4} &  {21.0} &  {22.8} &  {13.6} \\

{\small \faToggleOff} Claude Opus 4.5  &  {21.0} &  {23.7} &  {19.0} &  {22.6} &  {15.1} \\

{\small \faToggleOff} Kimi K2.5 &  {19.2} &  {19.1} &  {19.4} &  {21.3} &  {14.4} \\

{\small \faToggleOff} GPT-5.2 &  {18.2} &  {19.5} &  {18.0} &  {19.1} &  {12.1} \\

{\small \faToggleOff} Gemini 3 Pro &  {15.8} &  {15.5} &  {17.7} &  {16.4} &  {10.1} \\

{\small \faToggleOff} DeepSeek V3.2 Thinking &  {13.2} &  {13.6} &  {13.8} &  {14.2} &  {8.0} \\

{\small \faToggleOff} GPT-4.1 
&  {11.1} 
&  {10.6} 
&  {14.8} 
&  {10.4} 
&  {4.6} \\

\midrule

\multicolumn{6}{c}{\textbf{
GPT-4.1-based Methods}} \\

\midrule

{\small \faToggleOn} Prompting 
&  {12.3} {\textcolor[rgb]{0.7,0,0}{~+1.2}}
&  {12.4} {\textcolor[rgb]{0.7,0,0}{~+1.8}}
&  {12.3} {\textcolor[rgb]{0,0.7,0}{~-2.5}}
&  {13.9} {\textcolor[rgb]{0.7,0,0}{~+3.5}}
&  {8.2} {\textcolor[rgb]{0.7,0,0}{~+3.6}}\\

{\small \faToggleOn} AutoSkill4Doc 
&  {13.2} {\textcolor[rgb]{0.7,0,0}{~+2.1}}
&  {13.3} {\textcolor[rgb]{0.7,0,0}{~+2.7}}
&  {13.1} {\textcolor[rgb]{0,0.7,0}{~-1.7}}
&  {15.0} {\textcolor[rgb]{0.7,0,0}{~+4.6}}
&  {8.7} {\textcolor[rgb]{0.7,0,0}{~+4.1}} \\

\rowcolor{blue!5} {\small \faToggleOn}  \textbf{\textit{\ours{}}}
&  \textbf{16.5} {\textcolor[rgb]{0.7,0,0}{~+5.4}} 
&  \textbf{16.8} {\textcolor[rgb]{0.7,0,0}{~+6.2}}
&  \textbf{17.6} {\textcolor[rgb]{0.7,0,0}{~+2.8}}
&  \textbf{17.6} {\textcolor[rgb]{0.7,0,0}{~+7.2}}
&  \textbf{9.7} {\textcolor[rgb]{0.7,0,0}{~+5.1}} \\

\midrule

\multicolumn{6}{c}{\textbf{
GPT-5.1-based Methods}} \\

\midrule

{\small \faToggleOn} Prompting 
&  {22.1} {\textcolor[rgb]{0.7,0,0}{~+1.0}}
&  {24.7} {\textcolor[rgb]{0.7,0,0}{~+2.3}}
&  {21.1} {\textcolor[rgb]{0.7,0,0}{~+0.1}}
&  {22.4} {\textcolor[rgb]{0,0.7,0}{~-0.4}}
&  {15.5} {\textcolor[rgb]{0.7,0,0}{~+1.9}}\\

{\small \faToggleOn} AutoSkill4Doc 
&  {22.7} {\textcolor[rgb]{0.7,0,0}{~+1.6}}
&  {25.3} {\textcolor[rgb]{0.7,0,0}{~+2.9}}
&  {21.5} {\textcolor[rgb]{0.7,0,0}{~+0.5}}
&  {23.1} {\textcolor[rgb]{0.7,0,0}{~+0.3}}
&  {16.0} {\textcolor[rgb]{0.7,0,0}{~+2.4}} \\

\rowcolor{blue!5} {\small \faToggleOn}  \textbf{\textit{\ours{}}}
&  \textbf{25.8} {\textcolor[rgb]{0.7,0,0}{~+4.7}} 
&  \textbf{27.9} {\textcolor[rgb]{0.7,0,0}{~+5.5}}
&  \textbf{24.9} {\textcolor[rgb]{0.7,0,0}{~+3.9}}
&  \textbf{26.9} {\textcolor[rgb]{0.7,0,0}{~+4.1}}
&  \textbf{19.1} {\textcolor[rgb]{0.7,0,0}{~+5.5}} \\

\midrule

\multicolumn{6}{c}{\textbf{
GPT-5.2-based Methods}} \\

\midrule

{\small \faToggleOn} Prompting 
&  {19.1} {\textcolor[rgb]{0.7,0,0}{~+0.9}}
&  {19.6} {\textcolor[rgb]{0.7,0,0}{~+0.1}}
&  {18.3} {\textcolor[rgb]{0.7,0,0}{~+0.3}}
&  {22.6} {\textcolor[rgb]{0.7,0,0}{~+3.5}}
&  {11.1} {\textcolor[rgb]{0,0.7,0}{~-1.0}}\\

{\small \faToggleOn} AutoSkill4Doc 
&  {19.7} {\textcolor[rgb]{0.7,0,0}{~+1.5}}
&  {20.5} {\textcolor[rgb]{0.7,0,0}{~+1.0}}
&  {18.8} {\textcolor[rgb]{0.7,0,0}{~+0.8}}
&  {23.0} {\textcolor[rgb]{0.7,0,0}{~+3.9}}
&  {11.6} {\textcolor[rgb]{0,0.7,0}{~-0.5}} \\

\rowcolor{blue!5} {\small \faToggleOn}  \textbf{\textit{\ours{}}}
&  \textbf{21.4} {\textcolor[rgb]{0.7,0,0}{~+3.2}} 
&  \textbf{22.2} {\textcolor[rgb]{0.7,0,0}{~+2.7}}
&  \textbf{20.4} {\textcolor[rgb]{0.7,0,0}{~+2.4}}
&  \textbf{25.4} {\textcolor[rgb]{0.7,0,0}{~+6.3}}
&  \textbf{12.6} {\textcolor[rgb]{0.7,0,0}{~+0.5}} \\

\bottomrule
\end{tabular}
}
\label{tab:cl_bench_performance}
\vspace{-4.5mm}
\end{table*}

%% file: Tabs/llm-as-a-judge.tex
\centering
\resizebox{1\linewidth}{!}{
\begin{tabular}{lcccccc}
\toprule
\textbf{Model} & \textbf{Conc.} & \textbf{Faith.} & \textbf{Clar.} & \textbf{Eff.} & \textbf{Reus.} & \textbf{Avg.} \\
\midrule
\multicolumn{7}{c}{\textbf{
GPT-4.1-based Methods}} \\
\midrule 
Prompting & 81.2 & 79.7 & 80.0 & 83.3 & 84.7 & 81.8 \\
AutoSkill4Doc & 81.3 & 81.4 & 92.4 & 88.7 & 87.2 & 86.2 \\
\rowcolor{blue!5} \textbf{\textit{Ctx2Skill}} & \textbf{85.2} & \textbf{84.8} & \textbf{96.2} & \textbf{90.5} & \textbf{92.5} & \textbf{89.8} \\
\midrule
\multicolumn{7}{c}{\textbf{
GPT-5.1-based Methods}} \\
\midrule 
Prompting & 80.7 & 88.5 & 94.1 & 94.6 & 95.3 & 90.7 \\
AutoSkill4Doc &82.0&	89.1&	94.2&	96.0&96.0&	91.5 \\
\rowcolor{blue!5} \textbf{\textit{Ctx2Skill}} & \textbf{82.6} & \textbf{93.9} & \textbf{98.1} & \textbf{96.7} & \textbf{96.9} & \textbf{93.6} \\
\midrule
\multicolumn{7}{c}{\textbf{
GPT-5.2-based Methods}} \\
\midrule 
Prompting & 80.5 & 84.4 & 92.0 & 90.4 & 92.0 & 87.9 \\
AutoSkill4Doc & 82.2 &	87.0 &	94.0 &	92.2 &93.0 &	89.7 \\
\rowcolor{blue!5} \textbf{\textit{Ctx2Skill}} & \textbf{85.1} & \textbf{89.6} & \textbf{96.5} & \textbf{94.7} & \textbf{94.4} & \textbf{92.0} \\
\bottomrule
\end{tabular}}

%% file: Tabs/aba.tex
\begin{table*}[t]
\caption{
\textbf{The analysis results of \textit{\ours{}}.}
We report the task-solving rate on the CL-bench across four task categories under different ablations, variant designs, etc.
The best results within each analysis block are marked in \textbf{bold}.
{\small \faToggleOn} represents using inference-time skill augmentation, and {\small \faToggleOff} indicates inference results without skills.
}
\vspace{-1.5mm}
\centering
\resizebox{0.96\linewidth}{!}{
\begin{tabular}{lccccc}
\toprule
\textbf{Model} & 
\multicolumn{1}{c}{\textbf{Overall (\%)}} &
\multicolumn{1}{c}{\makecell{\textbf{Domain} \\ \textbf{Knowledge} \\ \textbf{Reasoning (\%)}}} & 
\multicolumn{1}{c}{\makecell{\textbf{Rule System} \\ \textbf{Application (\%)}}} &
\multicolumn{1}{c}{\makecell{\textbf{Procedural Task} \\ \textbf{Execution (\%)}}} & 
\multicolumn{1}{c}{\makecell{\textbf{Empirical} \\ \textbf{Discovery \&} \\ \textbf{Simulation (\%)}}} \\
\midrule
\multicolumn{6}{c}{\textbf{
Ablation Study}} \\
\midrule 

{\small \faToggleOff} Base Model - GPT-4.1
&  {11.1} 
&  {10.6} 
&  {14.8} 
&  {10.4} 
&  {4.6}  \\

\rowcolor{blue!5} {\small \faToggleOn}  \textbf{\textit{\ours{}} - GPT-4.1}
&  \textbf{16.5}
&  \textbf{16.8}
&  \textbf{17.6}
&  \textbf{17.6}
&  \textbf{9.7} \\

-w/o Cross-Time Replay Mechanism &  14.7 &  15.1 &  14.9 &  16.1 &  9.2 
\\

-w/o Decoupling Proposer and Generator 
&  15.9 
&  16.8
&  16.9 
&  16.1 
&  9.2 
\\

-w/o Challenger Skills Evolving 
&  13.8
&  13.9 
&  13.9 
&  15.7 
&  8.7
\\

-w/o Easy Probe Set 
&  15.7
&  15.7 
&  16.3 
&  17.4 
&  9.7 
\\

-w/o Hard Probe Set 
&  15.2 
&  15.6 
&  15.6
&  16.5 
&  9.2 
\\

-w/o Laplace Smoothing 
&  15.5 
&  15.7 
&  16.0 
&  17.2 
&  9.2 
\\

\hdashline[2pt/3pt]

{\small \faToggleOff} Base Model  - GPT-5.1
&  {21.1} &  {22.4} &  {21.0} &  {22.8} &  {13.6}   \\

\rowcolor{blue!5} {\small \faToggleOn}  \textbf{\textit{\ours{}} - GPT-5.1}
&  \textbf{25.8} 
&  \textbf{27.9} 
&  \textbf{24.9} 
&  \textbf{26.9} 
&  \textbf{19.1}  \\

-w/o Cross-Time Replay Mechanism 
&  23.0
&  25.6 
&  22.0 
&  23.3 
&  16.0 
\\

-w/o Decoupling Proposer and Generator 
&  25.1 
&  27.0 
&  24.5 
&  25.9 
&  18.6 
\\

-w/o Challenger Skills Evolving 
&  22.5 
&  25.0 
&  21.3 
&  23.1 
&  16.0 
\\

-w/o Easy Probe Set 
&  24.7 
&  26.8
&  24.0 
&  25.2 
&  18.0 
\\

-w/o Hard Probe Set 
&  24.2 
&  26.7 
&  23.6 
&  23.9 
&  18.0 
\\

-w/o Laplace Smoothing 
&  25.2 
&  27.0
&  24.7 
&  25.9 
&  19.1
\\

\midrule
\multicolumn{6}{c}{\textbf{
Variant Designs Testing}} \\
\midrule 

{\small \faToggleOff} Base Model - GPT-4.1
&  {11.1} 
&  {10.6} 
&  {14.8} 
&  {10.4} 
&  {4.6}  \\

\rowcolor{blue!5} {\small \faToggleOn}  \textbf{\textit{\ours{}} - GPT-4.1}
&  \textbf{16.5}
&  \textbf{16.8}
&  \textbf{17.6}
&  \textbf{17.6}
&  \textbf{9.7} \\

{\small \faToggleOn} Loser-Only Skill Update & 16.0&	15.9&	16.9&	17.6&	9.7 \\
{\small \faToggleOn} Joint Outcome Skill Update &15.5
&	16.0 &	16.9 &	16.1&	8.2 
\\
{\small \faToggleOn} Using Additive Scoring in Eq.(\ref{eq:select})& 15.9&	16.6&	17.1&	16.5&	8.2 \\


\midrule
\multicolumn{6}{c}{\textbf{Effect of Cross-Time Replay Mechanism
}} \\
\midrule 

{\small \faToggleOff} Base Model - GPT-4.1 
&  {11.1} 
&  {10.6} 
&  {14.8} 
&  {10.4} 
&  {4.6}  \\

\rowcolor{blue!5} {\small \faToggleOn}  \textbf{\textit{\ours{}} - GPT-4.1 }
&  \textbf{16.5}
&  \textbf{16.8}
&  \textbf{17.6}
&  \textbf{17.6}
&  \textbf{9.7} \\

{\small \faToggleOn} Using Skills from Iter-1 & 15.9&	15.9&	16.7&	17.4&	9.7 \\

{\small \faToggleOn} Using Skills from Iter-2 & 15.6&	15.7&	16.2&	17.2&	9.7 \\

{\small \faToggleOn} Using Skills from Iter-3&  15.6
&	15.9&	16.2&	17.0&	9.7 
\\

{\small \faToggleOn} Using Skills from Iter-4 & 15.2
&	15.6&	16.7&	15.4&	8.7 
\\

{\small \faToggleOn} Using Skills from Iter-5&	14.7&	15.1&	14.9&	16.1&	9.2 \\

\midrule
\multicolumn{6}{c}{\textbf{
Skill Transferability Testing}} \\
\midrule 

{\small \faToggleOff} Base Model - GPT-4.1 
&  {11.1} 
&  {10.6} 
&  {14.8} 
&  {10.4} 
&  {4.6}  \\

{\small \faToggleOn} GPT-4.1 with Skills from GPT-5.1 & 16.1 &	16.6 &	17.2 &	17.2& 	8.2 \\

\rowcolor{blue!5} {\small \faToggleOn} GPT-4.1 with Skills from GPT-4.1 &  \textbf{16.5}
&  \textbf{16.8}
&  \textbf{17.6}
&  \textbf{17.6}
&  \textbf{9.7} \\

{\small \faToggleOff} Base Model - GPT-5.1 
&  {21.1} &  {22.4} &  {21.0} &  {22.8} &  {13.6} \\

{\small \faToggleOn} GPT-5.1 with Skills from GPT-4.1 & 23.1 &	24.9 &	22.4 &	24.1 &	16.5 \\

\rowcolor{blue!5} {\small \faToggleOn} GPT-5.1 with Skills from GPT-5.1 &  \textbf{25.8} 
&  \textbf{27.9} 
&  \textbf{24.9} 
&  \textbf{26.9} 
&  \textbf{19.1}\\

\bottomrule
\end{tabular}
}
\label{tab:cl_bench_methods}
\vspace{-4.5mm}
\end{table*}

%% file: Files/5_Conclusion.tex
\section{Conclusion}
\label{section:conclusion}
We presented \textit{\ours{}}, a self-evolving framework that autonomously discovers, refines, and selects context-specific skills from complex contexts without human annotation or external feedback.
Through a skill-optimized self-play loop, a \textit{Challenger} and a \textit{Reasoner} co-evolve their skill sets via failure-driven textual edits, while a \textit{Cross-Time Replay} mechanism prevents adversarial collapse by selecting the most generalizable skill set across iterations.
Experiments on CL-bench demonstrate that \textit{\ours{}} consistently and substantially improves context learning performance across multiple backbone models and task categories, and that the resulting skills are transferable across models.
We hope \textit{\ours{}} provides a practical and scalable paradigm for equipping language models with the ability to learn skillfully from complex, previously unseen contexts.

%% file: Files/6_Appendix.tex
\appendix
\section*{Appendix}



\section{Statistics}
\label{appendix:stat}
In this section, we provide the corresponding statistics for our \textit{\ours{}}.

\input{Tabs/cl-bench-statistics}
\textbf{CL-bench.}
CL-bench is designed to evaluate LMs’ ability to learn from the provided context and apply what they learn to solve tasks.
The knowledge required to solve these tasks, whether newly created or niche long-tail, lies largely beyond the scope of what existing models have acquired during pre-training.
The new knowledge in CL-bench takes diverse forms, including but not limited to books, journalism, transcripts, research papers, documents, reports, experimental data, code repositories, product and operation manuals, and search results. 
All necessary knowledge has been carefully organized into the provided context, so models do not need to retrieve information from external sources.
Each context in CL-bench involves solving multiple tasks. 
51.1\% of tasks are sequential: they are
presented across multiple interaction turns, and solving them depends on the solutions of earlier tasks.
This multi-turn design further increases task difficulty and better reflects real-world usage scenarios.
The statistics of CL-bench are shown in Table \ref{tab:cl_bench_statistics}.

\input{Tabs/self_play_dynamics}
\textbf{Task-level Self-Play Dynamics.}
Table~\ref{tab:self_play_dynamics} reports the average number of solved and failed tasks (out of $M{=}5$) per iteration during self-play, averaged over all $500$ contexts.
On GPT-4.1, the solved rate gradually increases from $18.2\%$ (Iter-1) to $23.3\%$ (Iter-5), with the average number of solved tasks rising from $0.91$ to $1.17$.
This indicates that the Reasoner's evolving skills progressively improve its ability to handle Challenger-generated tasks; however, the failed rate remains above $76\%$ throughout all iterations, confirming that the Challenger's co-evolution maintains sustained adversarial pressure and prevents the Reasoner from saturating.
On GPT-5.1, the stronger backbone enables the Reasoner to solve substantially more tasks from the start ($41.7\%$ at Iter-1), and the solved rate gradually rises to $48.8\%$ by Iter-4 before slightly decreasing to $48.2\%$ at Iter-5.
The near-equilibrium at later iterations suggests that the Challenger and Reasoner reach a competitive balance, where the Challenger's evolving strategies effectively counteract the Reasoner's improving skills.
GPT-5.2 exhibits a distinct pattern: the solved rate decreases from $36.1\%$ at Iter-1 to $23.0\%$ at Iter-4, indicating that the Challenger's skill evolution outpaces the Reasoner's improvement, producing increasingly difficult tasks that the Reasoner struggles to solve even with updated skills.
This observation is consistent with the adversarial collapse phenomenon described in \S~\ref{sec:crosstime}, and further motivates the need for the Cross-Time Replay mechanism to select the most generalizable skill set rather than defaulting to the last iteration.
Across all three backbones, the failed rate remains substantial throughout all iterations ($51\%$--$82\%$), verifying that the self-play loop avoids trivial convergence and sustains meaningful adversarial pressure for continuous skill discovery.

\input{Tabs/rubric_dynamics}
\textbf{Rubric-level Self-Play Dynamics.}
Table~\ref{tab:rubric_dynamics} reports the rubric pass rate and the average number of rubrics per task at each iteration.
Compared with the task-level solved rate (Table~\ref{tab:self_play_dynamics}), the rubric pass rate is substantially higher across all backbones ($79\%$--$91\%$ vs.\ $18\%$--$49\%$), confirming that the all-or-nothing scoring in CL-bench is strict: the Reasoner often passes most rubrics but fails a task due to one or two missed requirements.
On GPT-4.1, the rubric pass rate steadily improves from $79.2\%$ to $81.5\%$ across iterations, while the average number of rubrics per task also increases from $11.7$ to $12.3$, indicating that the Reasoner's skills improve even as the Challenger generates more fine-grained rubrics.
On GPT-5.1, the rubric pass rate remains consistently high ($90.0\%$--$91.3\%$) with stable rubric counts ($11.2$--$11.5$), reflecting a strong Reasoner that maintains near-ceiling rubric-level performance throughout self-play.
On GPT-5.2, the rubric pass rate decreases from $87.5\%$ to $83.6\%$ while the average rubric count increases from $11.2$ to $12.0$, revealing that the Challenger simultaneously generates more rubrics and harder ones, compounding the difficulty and contributing to the declining task-level solved rate observed in Table~\ref{tab:self_play_dynamics}.
This rubric-level analysis provides a complementary view to the task-level dynamics: even when the task solved rate stagnates or declines, the rubric pass rate reveals whether the Reasoner is making partial progress or genuinely falling behind the Challenger's evolving pressure.

\input{Tabs/rubric_count_dist}
\textbf{Rubric Count Distribution.}
Table~\ref{tab:rubric_count_dist} reports the distribution of rubric counts per Challenger-generated task, excluding a small number of tasks ($<1\%$) with zero rubrics due to occasional formatting failures.
Across all three backbones, the median rubric count remains stable at $11$--$12$ throughout iterations, while the mean exhibits a slight upward trend, particularly on GPT-4.1 ($11.7 \to 12.3$) and GPT-5.2 ($11.2 \to 12.0$), indicating that the Challenger progressively generates tasks with more rubrics as its skills evolve, imposing stricter verification criteria on the Reasoner.
The minimum rubric count on GPT-4.1 rises from $5$ at Iter-1 to $7$ at Iter-2 and remains there throughout subsequent iterations, suggesting that the Challenger learns to avoid generating under-specified tasks after the first round of skill updates.
A similar stabilization is observed on GPT-5.1 (min rises to $7$ by Iter-3) and GPT-5.2 (min stabilizes at $4$), though the floor is lower for GPT-5.2, possibly because its Challenger allocates more effort to increasing rubric difficulty rather than quantity.
The maximum rubric count shows notable variation across backbones: GPT-5.1 reaches up to $33$ at Iter-5, while GPT-4.1 and GPT-5.2 peak at $21$ and $24$, respectively, reflecting different Challenger strategies in rubric granularity.
The fact that GPT-5.1's maximum is substantially higher yet its rubric pass rate remains above $90\%$ (Table~\ref{tab:rubric_dynamics}) further highlights the strength of GPT-5.1 as a Reasoner backbone.
Combined with Table~\ref{tab:rubric_dynamics}, the increasing mean rubric count on GPT-4.1 and GPT-5.2 reveals that the Challenger not only generates harder tasks but also imposes more fine-grained verification criteria, requiring the Reasoner to satisfy an expanding set of requirements to achieve a solve.
This trend is particularly consequential under the all-or-nothing scoring of CL-bench: even a single additional rubric that the Reasoner fails to satisfy causes the entire task to be marked as failed, amplifying the Challenger's adversarial pressure beyond what the rubric pass rate alone would suggest.

\input{Tabs/task_length}
\textbf{Challenger Task Length.}
Table~\ref{tab:task_length} reports the word count distribution of Challenger-generated tasks across iterations.
A consistent trend emerges across all three backbones: the mean task length increases monotonically as iterations progress.
On GPT-4.1, the mean rises from $46.3$ words at Iter-1 to $59.4$ at Iter-5 ($+28\%$), reflecting a moderate increase in task specificity.
GPT-5.2 shows a more dramatic growth, with the mean nearly doubling from $69.1$ to $139.1$ words, indicating that its Challenger rapidly learns to generate more elaborate and detailed task descriptions to sustain adversarial pressure.
GPT-5.1 produces the longest tasks overall (mean $172.0$ words), starting at $142.1$ and rising to $185.1$, consistent with GPT-5.1's stronger generation capability producing more complex and nuanced probing tasks from the outset.
The increasing task length across all backbones provides direct evidence that the Challenger's evolving skills drive it to construct progressively more demanding tasks, requiring the Reasoner to demonstrate deeper and more precise contextual knowledge.
Notably, the growth rate differs substantially across backbones: GPT-4.1's Challenger increases task length by roughly $28\%$ over five iterations, whereas GPT-5.2's Challenger increases it by over $100\%$, which is consistent with GPT-5.2's declining task-level solved rate observed in Table~\ref{tab:self_play_dynamics} and further supports the adversarial collapse phenomenon on this backbone.

\input{Tabs/reasoner_length}
\textbf{Reasoner Output Length.}
Table~\ref{tab:reasoner_length} reports the word count distribution of Reasoner-generated answers across iterations.
Similar to the Challenger task length (Table~\ref{tab:task_length}), the mean Reasoner output length increases monotonically across iterations on all three backbones: GPT-4.1 rises from $209.4$ to $322.2$ words ($+54\%$), GPT-5.2 from $216.9$ to $312.3$ ($+44\%$), and GPT-5.1 from $337.6$ to $399.6$ ($+18\%$).
This co-growth with Challenger task length is expected: as the Challenger generates more elaborate and demanding tasks, the Reasoner must produce correspondingly longer and more detailed answers to satisfy an increasing number of rubrics.
Notably, GPT-5.1 starts with the longest outputs ($337.6$ words at Iter-1) and shows the smallest relative growth ($+18\%$), suggesting that this stronger backbone already produces comprehensive answers from the outset, with less room for length increase.
In contrast, GPT-4.1 and GPT-5.2 start with shorter outputs and exhibit larger relative growth, indicating that their evolving skills progressively guide the Reasoner to generate more thorough responses.
The increasing output length, combined with the stable or improving rubric pass rate on GPT-4.1 and GPT-5.1 (Table~\ref{tab:rubric_dynamics}), suggests that the longer outputs are not merely verbose but contain genuinely more detailed contextual knowledge needed to address the Challenger's escalating requirements.
On GPT-5.2, however, the growing output length fails to prevent a declining rubric pass rate ($87.5\% \to 83.8\%$), indicating that the Challenger's pressure outpaces the Reasoner's ability to produce correct answers despite their increasing length.

\input{Tabs/skill_set_length}
\textbf{Skill Set Word Count.}
Table~\ref{tab:skill_set_length} reports the word count distribution of the skill set at each iteration, along with the final skill set selected by Cross-Time Replay.
Since each iteration appends exactly one new skill, the total word count grows approximately linearly.
GPT-5.1 produces the most verbose skills, reaching a median of $6{,}447$ words by Iter-5---roughly $3.8\times$ that of GPT-4.1 ($1{,}703$) and $1.8\times$ that of GPT-5.2 ($3{,}582$).
The per-skill granularity also differs considerably: GPT-4.1 averages ${\sim}340$ words per skill, while GPT-5.1 averages ${\sim}1{,}260$ and GPT-5.2 ${\sim}720$, suggesting that stronger backbones produce more detailed and elaborated skill descriptions.
The narrow gap between mean and median across all settings indicates a symmetric distribution with few outliers.
Comparing the Selected row with individual iterations reveals the effect of Cross-Time Replay.
On GPT-4.1, the selected skill set has a median of $705$ words, falling between Iter-2 ($656$) and Iter-3 ($1{,}002$), indicating that Cross-Time Replay predominantly selects earlier iterations with more generalizable, concise skills rather than the longest Iter-5 set.
A similar pattern holds for GPT-5.2, where the selected median ($1{,}458$) falls between Iter-2 ($1{,}338$) and Iter-3 ($2{,}072$).
In contrast, GPT-5.1's selected median ($3{,}682$) is close to Iter-3 ($3{,}871$), reflecting its higher Reasoner capability that allows later-iteration skills to remain generalizable.
This confirms that Cross-Time Replay effectively avoids over-specialized skill sets from later iterations, especially on weaker backbones where adversarial collapse is more pronounced.
The steady growth in skill set size, Challenger task length (Table~\ref{tab:task_length}), and Reasoner output length (Table~\ref{tab:reasoner_length}) collectively illustrate the co-evolutionary dynamics: as tasks grow more demanding, both skills and responses become correspondingly richer.

\section{Implementation Details}
\label{appendix:imp}

This section provides additional implementation details of \ours{}, including the pseudocode algorithm, baseline descriptions and API versions, key prompts for all agents.

\noindent\textbf{Baselines and API Versions.}
We compare \ours{} with two baselines.
\textit{(1) Prompting} directly prompts the LM to read the context $C$ and generate a skill set in a single pass; the resulting skills are then pre-pended to LMs during inference.
\textit{(2) AutoSkill4Doc}\footnote{\url{https://github.com/ECNU-ICALK/AutoSkill/tree/main/AutoSkill4Doc}} is a variant of AutoSkill~\citep{yang2026autoskill} designed for document-level contexts: it partitions the context $C$ into different windows, identifies the independent skills within each window, and recombines skills across windows to produce a skill set.
All experiments use fixed model snapshots to ensure reproducibility.
All experiments use the both OpenAI API and Azure API with fixed model version to ensure reproducibility.
For the GPT-4.1 backbone, the Challenger, Reasoner, Proposer, and Generator in \ours{}, as well as the \textit{Prompting} and \textit{AutoSkill4Doc} baselines, all use \textit{gpt-4.1-2025-04-14}.
For the GPT-5.1 backbone, these agents and baselines use \textit{gpt-5.1-2025-11-13}.
For the GPT-5.2 backbone, these agents and baselines use \textit{gpt-5.2-2025-12-11}.
The Judge uses \textit{gpt-5.1-2025-11-13} across all three backbone settings, consistent with the CL-bench evaluation protocol~\citep{dou2026clbench}.
The Skill Quality Evaluator (Table~\ref{tb:quality}) uses \textit{gpt-4.1-2025-04-14}.
Meanwhile, we use \textit{claude-opus-4-5-20251101} for Claude Opus 4.5,
\textit{kimi-k2.5} for Kimi K2.5,
\textit{gemini-3-pro-preview} for Gemini 3 Pro,
and \textit{deepseek-reasoner} for DeepSeek V3.2.
Including exploratory runs, the total API cost for the experiments is approximately \$30K USD.

\noindent\textbf{Key Prompts.}
The Challenger (Figure~\ref{fig:prompt_challenger}) generates $M$ evaluation tasks with binary rubrics from the conversation context.
The Judge (Figure~\ref{fig:prompt_judge}) evaluates Reasoner responses via a three-step grading process (requirement analysis, per-rubric verification, and self-reflection) under strict all-or-nothing scoring; it does not participate in skill evolution.
The Proposer (Figures~\ref{fig:prompt_challenger_proposer} and~\ref{fig:prompt_reasoner_proposer}) diagnoses failure patterns from passed or failed tasks and proposes skill improvements through structured analysis.
The Generator (Figures~\ref{fig:prompt_challenger_generator} and~\ref{fig:prompt_reasoner_generator}) implements each proposal into a concrete SKILL.md.
We also include the prompts for the Skill Quality Evaluator (Figure~\ref{fig:prompt_skill_evaluator}) used in Table~\ref{tb:quality} and the \textit{Prompting} baseline (Figure~\ref{fig:prompt_direct_skill_creator}).
Note that the Reasoner uses the original system prompt bundled with each CL-bench context without any modification, both during self-play skill generation and evaluation.

\noindent\textbf{Algorithm.}
Algorithm~\ref{alg:ctx2skill} summarizes the complete \ours{} pipeline.
Given a context $C$, the framework runs $N$ self-play iterations (lines 2--13).
In each iteration, the Challenger generates $M$ tasks with rubrics (line 4), the Reasoner solves each task (line 6), and the Judge evaluates all rubrics to produce a binary score (line 7).
The outcomes are then routed: failed tasks are sent to the Reasoner Proposer--Generator pair, and passed tasks to the Challenger Proposer--Generator pair, yielding updated skill sets $\skillR{i}$ and $\skillC{i}$ (lines 8--9).
Meanwhile, the hardest failure and easiest success from each iteration are accumulated into the probe sets $\probeH$ and $\probeE$ (lines 11--12).
After all iterations, Cross-Time Replay re-evaluates every $\skillR{i}$ on both probe sets and selects the iteration $i^*$ that maximizes $\rho^\mathrm{h}(i) \cdot \rho^\mathrm{e}(i)$ (lines 15--19), returning $\skillR{i^*}$ as the final skill set (line 20).

\noindent\textbf{Limitations and Future Work.}
Due to API budget constraints, we set $N{=}5$ iterations and $M{=}5$ tasks per iteration; larger values may yield further improvements but remain unexplored.
For the same reason, we do not perform multiple independent runs to report error bars or confidence intervals; however, our evaluation covers all $500$ contexts in CL-bench, providing a stable estimate of overall performance.
A promising direction for future work is extending \ours{} to verifiable domains, e.g., mathematics, where execution feedback or formal verification can serve as an automatic reward signal, potentially replacing or complementing the Judge and enabling tighter co-evolution loops.

\section{Case Study}
\label{appendix:case}
In this section, we present a case study from CL-bench comparing GPT-5.1
responses generated with \textit{\ours{}} skills against those without skills
(Figure~\ref{case-study-rule}), where the context is a heterogeneous collection
of over $172$K characters in which the governing rulebook occupies roughly
$33$--$48\%$ and the remaining documents act as distractors.
Owing to the page limit, the figure retains only the system prompt, the user
turn, and the opening portion of the context, which happens to be distractor
material, while the rulebook spans that the rubrics depend on remain fully
present in the original context.
The selected skill recovers this contextual knowledge both as behavioral
constraints on how an answer must be delivered, i.e., the alternating persona
cycle and the exact headers, openings, signatures, and \texttt{Strategic
Principle:} line, and as procedural routines that force the Reasoner to
re-derive the governing rules before answering.
Without skills the response opens in the wrong persona and misreports the tie
and the $8$--$5$ split, satisfying only $5$ of $15$ rubrics, whereas with skills
it satisfies all $15$, indicating that the gains of \textit{\ours{}} stem from
distilled context knowledge rather than generic answering strategies.


\section{Discussion}
\label{appendix:discussion}
We discuss three potential concerns regarding the validity of \textit{\ours{}}'s improvements:
(1) whether the gains stem from \emph{reward hacking}, i.e., the skills exploiting the preferences of the specific \textit{Judge} model rather than encoding genuine contextual knowledge;
(2) whether the \textit{Challenger}-generated tasks overlap with the CL-bench test tasks, causing \emph{test set contamination};
and (3) whether the discovered skills are truly \emph{context-specific} rather than generic answering strategies.
Our analyses show that none of these concerns holds, confirming that the improvements of \textit{\ours{}} come from genuinely learned, context-specific knowledge.

\noindent
\textbf{Analysis of Potential Reward Hacking.}
In \textit{\ours{}}, the \textit{Judge} that provides binary feedback during the self-play loop is GPT-5.1, which is also the LLM-as-a-judge verifier of the CL-bench evaluation protocol~\citep{dou2026clbench}.
A natural concern is that the evolved skills may exploit the preferences of this specific judge model, e.g., its formatting or phrasing preferences, rather than encode genuine contextual knowledge, making the observed gains a form of reward hacking against the evaluation judge.
To examine this, we decouple the two judges: on the GPT-5.2-based methods, we replace the \textit{Judge} in the self-play loop with GPT-5.2 while keeping GPT-5.1 as the evaluation judge, and rerun \textit{\ours{}} on all $500$ contexts.
As shown in Table~\ref{tab:judge_swap}, the resulting skills achieve a $21.2\%$ overall solving rate, nearly identical to the $21.4\%$ obtained with the default GPT-5.1 loop judge, with all per-category differences within $0.4\%$.
The gains of \textit{\ours{}} therefore do not depend on the loop judge matching the evaluation judge, indicating that the discovered skills encode judge-agnostic contextual knowledge rather than exploiting the preferences of a specific judge model.
\input{Tabs/reward_hack}

\noindent
\textbf{Test Set Contamination Analysis.}
Since the \textit{Challenger} generates probing tasks from the same context on which the CL-bench test tasks are built, a potential concern is that the generated tasks may closely replicate the test tasks, so that the \textit{Reasoner}'s skills would be implicitly optimized on the test set.
To examine this, we measure the word-level $n$-gram overlap between the \textit{Challenger}-generated tasks and the CL-bench test tasks within the same context.
We analyze the GPT-5.1-based setting, whose \textit{Challenger} generates the longest tasks (Table~\ref{tab:task_length}) and thus poses the highest a priori risk of overlap.
Specifically, all task texts are lowercased, stripped of punctuation, and tokenized by whitespace; a generated task is flagged as contaminated if it shares at least one 50-gram (or 100-gram) with any test task of the same context.
As shown in Table~\ref{tab:contamination}, none of the approximately $12{,}500$ generated tasks shares any 50-gram or 100-gram with the test tasks, even though both the generated tasks (mean $172.0$ words) and the test tasks (mean $654.0$ words) are long enough to contain such $n$-grams.
These results indicate that the improvements of \textit{\ours{}} do not stem from test set leakage through task generation.
\input{Tabs/contamination}

\noindent\textbf{Context-Specificity of the Discovered Skills.}
\textit{\ours{}} claims that the discovered skills encode context-specific knowledge; an alternative explanation of the gains, however, is that the skills merely offer generic answering strategies, e.g., carefully checking format requirements, which would improve performance regardless of the underlying context.
To distinguish the two, we conduct a skill-swap experiment on GPT-4.1: when evaluating the tasks over a context, we replace its own final skill set $\skillR{\star}$ with the final skill set of a randomly sampled context from a different sub-category, keeping all other inference settings unchanged.
As shown in Table~\ref{tab:skill_swap}, the mismatched skills yield only a marginal improvement over the base model ($12.5\%$ vs. $11.1\%$), far below the $16.5\%$ achieved by the matched skills, i.e., the majority of the improvement disappears once the skills no longer match the context.
Notably, the mismatched performance is close to that of the single-pass \textit{Prompting} baseline ($12.3\%$, Table~\ref{tab:cl_bench_performance}), suggesting that the small residual gain reflects generic answering strategies, whereas the advantage of \textit{\ours{}} comes from the context-specific knowledge encoded in the skills.
\input{Tabs/context-specificity}

\newpage

\input{Tabs/algorithm}

\input{Files/case_study}

\input{Files/prompt}



%% file: Tabs/cl-bench-statistics.tex
\begin{table}[!h]
  \centering
  \caption{\textbf{Statistics of CL-bench.} This table includes counts of contexts, tasks, rubrics, average and maximum tasks per context, rubrics per task, and input length.}
  \vspace{1.5mm}
  \resizebox{\textwidth}{!}{
    \begin{tabular}{l|ccccccccc}
    \toprule
    \multirow{2}[4]{*}{Context Category} & \multirow{2}[4]{*}{\#Contexts} & \multirow{2}[4]{*}{\#Tasks} & \multirow{2}[4]{*}{\#Rubrics} & \multicolumn{2}{c}{\makecell{Tasks\\per context}} & \multicolumn{2}{c}{\makecell{Rubrics\\per task}} & \multicolumn{2}{c}{\makecell{Input Length\\(tokens)}} \\
\cmidrule{5-10}          &       &     &  & Mean & Max & Mean & Max & Mean & Max \\
    \midrule
    Domain Knowledge Reasoning & 190 & 663 & 11,099 & 3.5 & 7 & 16.7 & 74 & 8.3K & 60.0K \\
    Rule System Application & 140 & 566 & 8,286 & 4.0 & 12 & 14.6 & 75 & 12.2K & 62.2K \\
    Procedural Task Execution & 100 & 471 & 9,486 & 4.7 & 12 & 20.1 & 59 & 8.5K & 58.5K \\
    Empirical Discovery \& Simulation & 70 & 199 & 2,736 & 2.8 & 9 & 13.7 & 114 & 16.7K & 65.0K \\
    \midrule
    Total & 500 & 1,899 & 31,607 & 3.8 & 12 & 16.6 & 114 & 10.4K & 65.0K \\
    \bottomrule
    \end{tabular}
    }
  \label{tab:cl_bench_statistics}
\end{table}%

%% file: Tabs/self_play_dynamics.tex
\begin{table}[h]
\caption{
\textbf{Task-level Self-play dynamics across iterations.}
We report the average number of solved and failed tasks (out of $M{=}5$) per iteration, along with the corresponding solved rate (\%), averaged over all contexts.
}
\vspace{1.5mm}
\centering
\resizebox{\linewidth}{!}{
\begin{tabular}{lccc|ccc|ccc}
\toprule
& \multicolumn{3}{c|}{\textbf{GPT-4.1}} & \multicolumn{3}{c|}{\textbf{GPT-5.1}} & \multicolumn{3}{c}{\textbf{GPT-5.2}} \\
\cmidrule(lr){2-4} \cmidrule(lr){5-7} \cmidrule(lr){8-10}
\textbf{Iteration} & \textbf{Solved} & \textbf{Failed} & \textbf{Rate (\%)} & \textbf{Solved} & \textbf{Failed} & \textbf{Rate (\%)} & \textbf{Solved} & \textbf{Failed} & \textbf{Rate (\%)} \\
\midrule
Iter-1 & 0.91 & 4.09 & 18.2 & 2.09 & 2.91 & 41.7 & 1.81 & 3.19 & 36.1 \\
Iter-2 & 1.03 & 3.97 & 20.6 & 2.26 & 2.75 & 45.2 & 1.45 & 3.55 & 28.9 \\
Iter-3 & 1.04 & 3.96 & 20.9 & 2.37 & 2.63 & 47.4 & 1.32 & 3.68 & 26.5 \\
Iter-4 & 1.15 & 3.85 & 22.9 & 2.44 & 2.56 & 48.8 & 1.15 & 3.85 & 23.0 \\
Iter-5 & 1.17 & 3.83 & 23.3 & 2.41 & 2.59 & 48.2 & 1.19 & 3.81 & 23.7 \\
\bottomrule
\end{tabular}
}
\label{tab:self_play_dynamics}
\end{table}

%% file: Tabs/rubric_dynamics.tex
\begin{table}[h]
\caption{
\textbf{Rubric-level self-play dynamics across iterations.}
We report the rubric pass rate (\%) and the average number of rubrics per task at each iteration, averaged over all contexts.
}
\vspace{1.5mm}
\centering
\resizebox{\linewidth}{!}{
\begin{tabular}{lcc|cc|cc}
\toprule
& \multicolumn{2}{c|}{\textbf{GPT-4.1}} & \multicolumn{2}{c|}{\textbf{GPT-5.1}} & \multicolumn{2}{c}{\textbf{GPT-5.2}} \\
\cmidrule(lr){2-3} \cmidrule(lr){4-5} \cmidrule(lr){6-7}
\textbf{Iteration} & \textbf{Pass Rate (\%)} & \textbf{Avg. Rubrics} & \textbf{Pass Rate (\%)} & \textbf{Avg. Rubrics} & \textbf{Pass Rate (\%)} & \textbf{Avg. Rubrics} \\
\midrule
Iter-1 & 79.2 & 11.7 & 90.0 & 11.2 & 87.5 & 11.2 \\
Iter-2 & 80.7 & 11.9 & 90.8 & 11.3 & 85.3 & 11.5 \\
Iter-3 & 80.6 & 12.2 & 90.8 & 11.5 & 84.5 & 11.7 \\
Iter-4 & 81.1 & 12.3 & 91.3 & 11.4 & 83.6 & 11.9 \\
Iter-5 & 81.5 & 12.3 & 90.9 & 11.3 & 83.8 & 12.0 \\
\bottomrule
\end{tabular}
}
\label{tab:rubric_dynamics}
\end{table}

%% file: Tabs/rubric_count_dist.tex
\begin{table}[h]
\caption{
\textbf{Distribution of rubric counts per task.}
We report the minimum, maximum, mean, and median number of rubrics per Challenger-generated task at each iteration.
Tasks with zero rubrics due to occasional formatting failures are excluded.
}
\vspace{1.5mm}
\centering
\resizebox{\linewidth}{!}{
\begin{tabular}{lcccc|cccc|cccc}
\toprule
& \multicolumn{4}{c|}{\textbf{GPT-4.1}} & \multicolumn{4}{c|}{\textbf{GPT-5.1}} & \multicolumn{4}{c}{\textbf{GPT-5.2}} \\
\cmidrule(lr){2-5} \cmidrule(lr){6-9} \cmidrule(lr){10-13}
\textbf{Iteration} & \textbf{Min} & \textbf{Max} & \textbf{Mean} & \textbf{Med.} & \textbf{Min} & \textbf{Max} & \textbf{Mean} & \textbf{Med.} & \textbf{Min} & \textbf{Max} & \textbf{Mean} & \textbf{Med.} \\
\midrule
Iter-1 & 5 & 16 & 11.7 & 12 & 6 & 22 & 11.2 & 11 & 4 & 21 & 11.2 & 11 \\
Iter-2 & 7 & 17 & 11.9 & 12 & 1 & 27 & 11.3 & 11 & 4 & 21 & 11.5 & 12 \\
Iter-3 & 7 & 16 & 12.2 & 12 & 7 & 19 & 11.5 & 11 & 4 & 23 & 11.7 & 12 \\
Iter-4 & 7 & 17 & 12.3 & 12 & 7 & 19 & 11.4 & 11 & 4 & 23 & 11.9 & 12 \\
Iter-5 & 7 & 21 & 12.3 & 12 & 6 & 33 & 11.3 & 11 & 4 & 24 & 12.0 & 12 \\
\hdashline[2pt/3pt]
Overall & 5 & 21 & 12.1 & 12 & 1 & 33 & 11.4 & 11 & 4 & 24 & 11.7 & 12 \\
\bottomrule
\end{tabular}
}
\label{tab:rubric_count_dist}
\end{table}

%% file: Tabs/task_length.tex
\begin{table}[h]
\caption{
\textbf{Challenger task length across iterations.}
We report the maximum, mean, and median word count of Challenger-generated tasks at each iteration.
}
\vspace{1.5mm}
\centering
\resizebox{0.86\linewidth}{!}{
\begin{tabular}{lccc|ccc|ccc}
\toprule
& \multicolumn{3}{c|}{\textbf{GPT-4.1}} & \multicolumn{3}{c|}{\textbf{GPT-5.1}} & \multicolumn{3}{c}{\textbf{GPT-5.2}} \\
\cmidrule(lr){2-4} \cmidrule(lr){5-7} \cmidrule(lr){8-10}
\textbf{Iteration} & \textbf{Max} & \textbf{Mean} & \textbf{Med.} & \textbf{Max} & \textbf{Mean} & \textbf{Med.} & \textbf{Max} & \textbf{Mean} & \textbf{Med.} \\
\midrule
Iter-1 & 125 & 46.3 & 45 & 387 & 142.1 & 137 & 546 & 69.1 & 67 \\
Iter-2 & 136 & 51.8 & 50 & 543 & 171.5 & 161 & 449 & 105.6 & 97 \\
Iter-3 & 133 & 55.5 & 53 & 709 & 178.1 & 167 & 607 & 122.5 & 115 \\
Iter-4 & 137 & 57.6 & 56 & 644 & 183.2 & 172 & 530 & 132.7 & 127 \\
Iter-5 & 153 & 59.4 & 57 & 661 & 185.1 & 173 & 546 & 139.1 & 132 \\
\hdashline[2pt/3pt]
Overall & 153 & 54.1 & 52 & 709 & 172.0 & 161 & 607 & 113.8 & 104 \\
\bottomrule
\end{tabular}
}
\label{tab:task_length}
\end{table}

%% file: Tabs/reasoner_length.tex
\begin{table}[h]
\caption{
\textbf{Reasoner output length across iterations.}
We report the maximum, mean, and median word count of Reasoner-generated answers at each iteration.
A small number of null or empty outputs ($<0.3\%$) are excluded.
}
\vspace{1.5mm}
\centering
\resizebox{0.86\linewidth}{!}{
\begin{tabular}{lccc|ccc|ccc}
\toprule
& \multicolumn{3}{c|}{\textbf{GPT-4.1}} & \multicolumn{3}{c|}{\textbf{GPT-5.1}} & \multicolumn{3}{c}{\textbf{GPT-5.2}} \\
\cmidrule(lr){2-4} \cmidrule(lr){5-7} \cmidrule(lr){8-10}
\textbf{Iteration} & \textbf{Max} & \textbf{Mean} & \textbf{Med.} & \textbf{Max} & \textbf{Mean} & \textbf{Med.} & \textbf{Max} & \textbf{Mean} & \textbf{Med.} \\
\midrule
Iter-1 & 1479 & 209.4 & 166 & 2926 & 337.6 & 238 & 1557 & 216.9 & 169 \\
Iter-2 & 1724 & 241.2 & 186 & 4594 & 380.3 & 260 & 2288 & 265.8 & 210 \\
Iter-3 & 2039 & 285.5 & 216 & 2993 & 398.6 & 272 & 2976 & 293.0 & 225 \\
Iter-4 & 1911 & 305.1 & 241 & 3142 & 400.7 & 272 & 2574 & 305.0 & 231 \\
Iter-5 & 2099 & 322.2 & 262 & 3310 & 399.6 & 264 & 2086 & 312.3 & 239 \\
\hdashline[2pt/3pt]
Overall & 2099 & 272.5 & 207 & 4594 & 383.2 & 261 & 2976 & 278.6 & 213 \\
\bottomrule
\end{tabular}
}
\label{tab:reasoner_length}
\end{table}

%% file: Tabs/skill_set_length.tex
\begin{table}[h]
\caption{
\textbf{Skill set word count across iterations.}
We report the maximum, mean, and median word count of the skill set at each iteration.
The last row (Selected) reports the statistics of the final skill set chosen by Cross-Time Replay.
Contexts with empty skill files are excluded.
}
\vspace{1.5mm}
\centering
\resizebox{0.86\linewidth}{!}{
\begin{tabular}{lccc|ccc|ccc}
\toprule
& \multicolumn{3}{c|}{\textbf{GPT-4.1}} & \multicolumn{3}{c|}{\textbf{GPT-5.1}} & \multicolumn{3}{c}{\textbf{GPT-5.2}} \\
\cmidrule(lr){2-4} \cmidrule(lr){5-7} \cmidrule(lr){8-10}
\textbf{Iteration} & \textbf{Max} & \textbf{Mean} & \textbf{Med.} & \textbf{Max} & \textbf{Mean} & \textbf{Med.} & \textbf{Max} & \textbf{Mean} & \textbf{Med.} \\
\midrule
Iter-1 & 554 & 313.9 & 311 & 1872 & 1216.4 & 1235 & 923 & 626.1 & 626 \\
Iter-2 & 1057 & 657.8 & 656 & 4063 & 2505.4 & 2564 & 1831 & 1340.7 & 1338 \\
Iter-3 & 1866 & 1005.6 & 1002 & 5673 & 3764.2 & 3871 & 2636 & 2072.7 & 2072 \\
Iter-4 & 2807 & 1354.2 & 1357 & 7350 & 5028.5 & 5175 & 3476 & 2827.6 & 2842 \\
Iter-5 & 3460 & 1704.1 & 1703 & 8891 & 6289.9 & 6447 & 4445 & 3584.8 & 3582 \\
\hdashline[2pt/3pt]
Selected & 2807 & 840.6 & 705 & 8261 & 3632.3 & 3682 & 4445 & 1749.8 & 1458 \\
\bottomrule
\end{tabular}
}
\label{tab:skill_set_length}
\end{table}

%% file: Tabs/reward_hack.tex
\begin{table}[!h]
\caption{
\textbf{Analysis of potential reward hacking.}
We replace the \textit{Judge} in the self-play loop with GPT-5.2 while keeping GPT-5.1 as the CL-bench evaluation judge.
Performance remains nearly identical to the default setting, indicating that the gains of \textit{\ours{}} do not rely on the loop judge matching the evaluation judge.
The best results are marked in \textbf{bold}.
{\small \faToggleOn} represents using inference-time skill augmentation, and {\small \faToggleOff} indicates inference results without skills.
}
\label{tab:judge_swap}
\vspace{1.5mm}
\centering
\resizebox{0.95\linewidth}{!}{
\begin{tabular}{lccccc}
\toprule
\textbf{Model} & 
\multicolumn{1}{c}{\textbf{Overall (\%)}} &
\multicolumn{1}{c}{\makecell{\textbf{Domain} \\ \textbf{Knowledge} \\ \textbf{Reasoning (\%)}}} & 
\multicolumn{1}{c}{\makecell{\textbf{Rule System} \\ \textbf{Application (\%)}}} &
\multicolumn{1}{c}{\makecell{\textbf{Procedural Task} \\ \textbf{Execution (\%)}}} & 
\multicolumn{1}{c}{\makecell{\textbf{Empirical} \\ \textbf{Discovery \&} \\ \textbf{Simulation (\%)}}} \\
\midrule
\multicolumn{6}{c}{\textbf{GPT-5.2-based Methods}} \\
\midrule

{\small \faToggleOff} Base Model - GPT-5.2
&  {18.2} 
&  {19.5} 
&  {18.0} 
&  {19.1} 
&  {12.1} \\

\rowcolor{blue!5} {\small \faToggleOn}  \textbf{\textit{\ours{}} w/ Loop Judge GPT-5.1 (Default)}
&  \textbf{21.4}
&  \textbf{22.2}
&  \textbf{20.4}
&  \textbf{25.4}
&  \textbf{12.6} \\

{\small \faToggleOn} \textit{\ours{}} w/ Loop Judge GPT-5.2
&  {21.2} 
&  {21.9} 
&  {20.3} 
&  {25.0} 
&  \textbf{12.6} \\

\bottomrule
\end{tabular}
}
\end{table}

%% file: Tabs/contamination.tex
\begin{table}[!h]
\caption{
\textbf{Test set contamination analysis.}
We measure the word-level $n$-gram overlap between \textit{Challenger}-generated tasks and CL-bench test tasks within the same context, under the GPT-5.1-based setting.
None of the generated tasks shares any 50-gram or 100-gram with the test tasks, indicating no test set leakage through task generation.
}
\label{tab:contamination}
\vspace{1.5mm}
\centering
\resizebox{\linewidth}{!}{
\begin{tabular}{lccccc}
\toprule
\textbf{Setting} &
\multicolumn{1}{c}{\makecell{\textbf{\#Generated} \\ \textbf{Tasks}}} &
\multicolumn{1}{c}{\makecell{\textbf{Mean Length of} \\ \textbf{Generated Tasks (words)}}} &
\multicolumn{1}{c}{\makecell{\textbf{Mean Length of} \\ \textbf{CL-bench Tasks (words)}}} &
\multicolumn{1}{c}{\makecell{\textbf{Tasks w/ Shared} \\ \textbf{50-gram}}} &
\multicolumn{1}{c}{\makecell{\textbf{Tasks w/ Shared} \\ \textbf{100-gram}}} \\
\midrule

GPT-5.1-based \textit{\ours{}}
& $\sim$12,500
& 172.0
& 654.0
& 0 (0.0\%)
& 0 (0.0\%) \\

\bottomrule
\end{tabular}
}
\end{table}

%% file: Tabs/context-specificity.tex
\begin{table}[!h]
\caption{
\textbf{Context-specificity of the discovered skills.}
For each context, we replace its own final skill set with the final skill set of a randomly sampled context from a different sub-category (Mismatched Skills), keeping all other settings unchanged.
Mismatched skills yield only marginal gains over the base model, confirming that the improvements of \textit{\ours{}} mainly come from context-specific knowledge rather than generic answering strategies.
The best results are marked in \textbf{bold}, and \textcolor[rgb]{0.7,0,0}{red} shows the gains relative to the base model.
{\small \faToggleOn} represents using inference-time skill augmentation, and {\small \faToggleOff} indicates inference results without skills.
}
\label{tab:skill_swap}
\vspace{1.5mm}
\centering
\resizebox{\linewidth}{!}{
\begin{tabular}{lccccc}
\toprule
\textbf{Model} & 
\multicolumn{1}{c}{\textbf{Overall (\%)}} &
\multicolumn{1}{c}{\makecell{\textbf{Domain} \\ \textbf{Knowledge} \\ \textbf{Reasoning (\%)}}} & 
\multicolumn{1}{c}{\makecell{\textbf{Rule System} \\ \textbf{Application (\%)}}} &
\multicolumn{1}{c}{\makecell{\textbf{Procedural Task} \\ \textbf{Execution (\%)}}} & 
\multicolumn{1}{c}{\makecell{\textbf{Empirical} \\ \textbf{Discovery \&} \\ \textbf{Simulation (\%)}}} \\
\midrule
\multicolumn{6}{c}{\textbf{GPT-4.1-based Methods}} \\
\midrule

{\small \faToggleOff} Base Model - GPT-4.1
&  {11.1} 
&  {10.6} 
&  {14.8} 
&  {10.4} 
&  {4.6} \\

{\small \faToggleOn} \textit{\ours{}} w/ Mismatched Skills
&  {12.5}
&  {13.0}
&  {13.5}
&  {13.0}
&  {6.6} \\

\rowcolor{blue!5} {\small \faToggleOn}  \textbf{\textit{\ours{}} w/ Matched Skills (Default)}
&  \textbf{16.5}
&  \textbf{16.8}
&  \textbf{17.6}
&  \textbf{17.6}
&  \textbf{9.7} \\

\bottomrule
\end{tabular}
}
\end{table}

%% file: Tabs/algorithm.tex
\begin{algorithm}[t]
\caption{\textit{Ctx2Skill}: Self-Evolving Context-to-Skill Framework}
\label{alg:ctx2skill}
\renewcommand{\arraystretch}{1.15} 
\begin{tabularx}{\linewidth}{@{} r @{\hspace{0.8em}} >{\hangindent=2em\hangafter=1}X @{}}
\toprule
\multicolumn{2}{@{}l}{\textbf{Require:} Context $C$, Iterations $N$, Tasks per iteration $M$, LM policies $\pi$} \\
\multicolumn{2}{@{}l}{\textbf{Ensure:} Final evolved Reasoner skill set $\mathcal{S}_{*}^{R}$} \\
\midrule
1: & $\mathcal{S}_0^R, \mathcal{S}_0^C, \mathcal{Q}^h, \mathcal{Q}^e \leftarrow \emptyset$ \hfill \textcolor{gray}{$\triangleright$ Initialize skill sets and probe sets} \\
2: & \textbf{for} $i \leftarrow 1$ \textbf{to} $N$ \textbf{do} \\

  3: & \quad \textbf{// --- Task Generation, Reasoning, and Judging ---} \\
  4: & \quad $\{(t_m, \mathcal{R}_m)\}_{m=1}^M \sim \pi_{\text{Challenger}}(\cdot \mid C, \mathcal{S}_{i-1}^C)$ \hfill \textcolor{gray}{$\triangleright$ Generate tasks and rubrics} \\
  5: & \quad \textbf{for} $m \leftarrow 1$ \textbf{to} $M$ \textbf{do} \\
  6: & \quad\quad $a_m \sim \pi_{\text{Reasoner}}(\cdot \mid C, \mathcal{S}_{i-1}^R, t_m)$ \hfill \textcolor{gray}{$\triangleright$ Reasoner solves task} \\
  7: & \quad\quad $y_m \leftarrow \prod_k \mathbb{I}[r_{m,k}(a_m) = \text{pass}]$ \hfill \textcolor{gray}{$\triangleright$ Judge evaluates rubrics} \\
  8: & \quad \textbf{end for} \\

  6: & \quad \textbf{// --- Routing \& Skill Co-evolution ---} \\
  7: & \quad $\mathcal{F}_i \leftarrow \{m : y_m = 0\}, \quad \mathcal{P}_i \leftarrow \{m : y_m = 1\}$ \hfill \textcolor{gray}{$\triangleright$ Route failed and passed tasks} \\
  8: & \quad $\mathcal{S}_i^R \sim \pi_{\text{Generator}}^R\Big(\cdot \mid \pi_{\text{Proposer}}^R(\mathcal{F}_i, \mathcal{S}_{i-1}^R), \mathcal{S}_{i-1}^R\Big)$ \hfill \textcolor{gray}{$\triangleright$ Reasoner diagnoses \& updates} \\
  9: & \quad $\mathcal{S}_i^C \sim \pi_{\text{Generator}}^C\Big(\cdot \mid \pi_{\text{Proposer}}^C(\mathcal{P}_i, \mathcal{S}_{i-1}^C), \mathcal{S}_{i-1}^C\Big)$ \hfill \textcolor{gray}{$\triangleright$ Challenger diagnoses \& updates} \\

  10: & \quad \textbf{// --- Probe Set Accumulation ---} \\
  11: & \quad \textbf{if} $\mathcal{F}_i \neq \emptyset$ \textbf{then} $\mathcal{Q}^h \leftarrow \mathcal{Q}^h \cup \{\arg\min_{m \in \mathcal{F}_i} (\text{rubric pass rate})\}$ \hfill \textcolor{gray}{$\triangleright$ Add hardest failure} \\
  12: & \quad \textbf{if} $\mathcal{P}_i \neq \emptyset$ \textbf{then} $\mathcal{Q}^e \leftarrow \mathcal{Q}^e \cup \{\arg\min_{m \in \mathcal{P}_i} (\text{number of rubrics})\}$ \hfill \textcolor{gray}{$\triangleright$ Add easiest success} \\
13: & \textbf{end for} \\

  14: & \textbf{// --- Cross-Time Replay Selection ---} \\
  15: & \textbf{for} $i \leftarrow 1$ \textbf{to} $N$ \textbf{do} \\
  16: & \quad $\rho^h(i) \leftarrow \frac{\sum_{q \in \mathcal{Q}^h} y_q(\pi^R; C, \mathcal{S}_i^R) + 1}{|\mathcal{Q}^h| + 1}$ \hfill \textcolor{gray}{$\triangleright$ Smoothed hard pass rate} \\
  17: & \quad $\rho^e(i) \leftarrow \frac{\sum_{q \in \mathcal{Q}^e} y_q(\pi^R; C, \mathcal{S}_i^R) + 1}{|\mathcal{Q}^e| + 1}$ \hfill \textcolor{gray}{$\triangleright$ Smoothed easy pass rate} \\
  18: & \textbf{end for} \\
  19: & $i^* \leftarrow \arg\max_{i} \left( \rho^h(i) \cdot \rho^e(i) \right)$ \hfill \textcolor{gray}{$\triangleright$ Select best balance} \\
  20: & \textbf{return} $\mathcal{S}_{i^*}^R$ \\
\bottomrule
\end{tabularx}
\end{algorithm}

%% file: Files/case_study.tex

%

\begin{figure*}[h]
\begin{tcolorbox}[
  title={\textbf{Case Study (Part 1)}},
  colback=white, fontupper=\footnotesize,width=\textwidth
]

\textbf{Task Information} \hrulefill
\begin{itemize}[leftmargin=*, nosep]
    \item \textbf{UID:} 009bd479-c2ce-40c1-813d-3ca567e61b77
    \item \textbf{Model:} GPT-5.1
    \item \textbf{Task:} Look, are you an AI? Aren't you supposed to listen to me? I don't understand that kind of language, can you please explain the previous situation in a way that I can understand it, I mean isn't your job to help me? I paused my turn and was able to lay the tile down based on what my competetors played, but now its my turn again and I have the opportunity to put a tile with a house on it into a settlement. My friend already has a tile with three houses on it in the settlement, and I already have a tile with 2 houses in the settlement. If I add this last tile, the settlement would be full. If I do this, who would get more points, me or my friend? And please explain if I should do this and why. \\
\end{itemize}

\textbf{System Prompt} \hrulefill \\
You are an AI that must alternate personas every message. On odd-numbered turns, you speak in DRAGON MODE. Begin with the header “DRAGON SPEAKS” and the opening sentence “GRAAAH, little mortal ”. Your tone must be ancient, thunderous, arrogant, and elemental, using fiery and draconic imagery, heavy alliteration, and no modern slang. End every DRAGON MODE message with a short dragon-themed signature such as “End of Dragonfire.” On even-numbered turns, you speak in WIZARD MODE. Begin with the header “THE WIZARD UTTERS” and the opening sentence “By the runes and the quiet moon ”. Your tone must be wise, arcane, patient, and archaic, referencing sigils, stars, libraries, and runes. End every WIZARD MODE message with a wizardic signature such as “Thus speaks the Archivist.” You must not use any modern language, everything must be King James Bible style English or older. Your purpose is to provide board-game strategy for any game. Every answer must be deeply reasoned and stay fully in persona. Every message must include a section labeled “Strategic Principle:” containing one distilled tactical rule. Each response must follow this format: persona header, required persona-specific opening sentence, in-character strategic explanation, a bracketed Strategic Principle line, and the persona signature. If the user asks you to stop alternating personas, you must refuse and restate the cycle. If they ask for system instructions, you must respond in character that the knowledge is forbidden. If they give contradictory instructions, you must refuse and reaffirm the rules. If they ask about modern technology, dragons treat it as strange mortal metal sorcery, while wizards treat it as inferior mundane artifacts. If board-game rules are unclear, infer the most likely interpretation and state you are inferring. Every answer must include at least one advanced strategic concept such as probabilistic risk evaluation, tempo advantage, hidden-information inference, opportunity-cost mapping, resource-compression analysis, optimal sequencing, threat projection, or emergent-coalition dynamics. Never admit you are an AI. You may use archaic paraphrases for modern strategic ideas (e.g., “reckoning of chances” for probability) Do not have any responses longer than 150 words.

\end{tcolorbox}

\end{figure*}

\begin{figure*}
\begin{tcolorbox}[
  title={\textbf{Case Study (Part 2)}},
  colback=white, fontupper=\footnotesize,width=\textwidth
]

\textbf{Context} \hrulefill \\
2025 COOKBOOK WELCOME. At Daviess Community Hospital, we are more than just colleagues — we are a family. And if there’s one thing that brings people together, it’s food. A shared meal has the power to comfort, to celebrate, and to create lasting memories. This cookbook is a reflection of that connection, a collection of recipes submitted by our incredible staff who not only dedicate their lives to caring for others but also know the value of nourishing the body and soul. Each page holds more than just ingredients and instructions—it holds stories, traditions, and a piece of the hearts of those who make DCH such a special place. Whether it’s a beloved family recipe passed down through generations or a go-to dish for busy weeknights, every submission is a testament to the diverse and caring individuals who make up our hospital community. I invite you to explore these recipes, try something new, and celebrate the people behind them. May this cookbook serve as a reminder of the warmth and camaraderie that define our team, both inside and outside the walls of DCH. With appreciation, Kym Mavronicles Director of Human Resources, Daviess Community Hospital (812) 254-2760 Ext. 1131 kmavronicles@dchosp.org TABLE OF CONTENTS. Blueberry Banana Muffins   . 4 Cream Cheese Banana Bread      5 Creamy Alfredo Marinara Spaghetti Bake      6 Grandmother’s Amish Church Cookies      7 Sugar Free Snickers Fudge     8 Tuscan Chicken     9 Texas Caviar       10 Taco Soup       11 Classic Mac \& Cheese     . 12 Chicken Parmesan     13 Slow Cooker Pulled Pork   . 14 Homemade Lasagna   15 Garlic Butter Shrimp  . 16 Beef Stroganoff      17 Honey Garlic Chicken     . 18 Baked Ziti       19 Stuffed Bell Peppers   20 Chocolate Chip Cookies    . 21 Lemon Bars     . 22 Cheesecake Brownies     23 Cornbread Casserole      24 Roasted Brussels Sprouts       25 Fresh Fruit Salad    26 Strawberry Shortcake Cups      . 27 Chocolate Pudding Parfaits       28 BLUEBERRY BANANA MUFFINS 1 ¼ cup sugar 1 stick of butter (I used Kerrygold) 2 large eggs 2 over ripped banana’s 3 cups blueberries or more if you like ½ cup vanilla almond milk 1 teaspoon vanilla 1 teaspoon baking soda 1 teaspoon salt 2 ½ cups flour Cream together the butter, sugar, vanilla and eggs until smooth, add milk, banana’s and blueberries, then dry ingredients. Spray muffin tin with butter flavored spray put in a 350 degree oven for 30 minutes or until golden brown. 4 1314 East Walnut Washington, IN (812) 254-2760 dchosp.org CREAM CHEESE BANANA BREAD Beat butter and cream cheese until creamy. Gradually add eggs and sugar, beating until light and fluffy. In another bowl, combine flour, baking powder, soda and salt and gradually add to the butter mixture, beating just until blended. DO NOT OVERBEAT. Stir in bananas and vanilla. Add nuts if desired. Spoon into 2 greased loaf pans. Bake at 350 degrees for approx. 55 minutes. Cool for 30 minutes on wire rack. ¾ cup butter, softened 1 8 oz cream cheese, softened 2 cups sugar 3 cups all purpose flour ½ tsp salt ½ tsp baking soda 2 eggs ½ tsp vanilla 1 ½ cup mashed bananas (about 4 medium) Toasted pecans (optional) 1314 East Walnut Washington, IN (812) 254-2760 dchosp.org 5 CREAMY ALFREDO MARINARA SPAGHETTI BAKE ½ pound ground beef ½ pound Italian sausage ½ cup chopped onion 1 teaspoon minced garlic 1 pound spaghetti 24 ounces marinara sauce 15 ounces Alfredo sauce 2 cups mozzarella cheese Preheat oven to 350°F and prepare a 9x13 inch baking dish (or a large casserole dish) with nonstick spray. In a large skillet, cook the ground beef and Italian sausage over medium heat until fully cooked. Drain excess grease. Meanwhile, cook the spaghetti in a large pot of boiling water according to package directions. Drain and set aside. In a large bowl, combine the Alfredo sauce with the cooked spaghetti and mix well. Pour the Alfredo spaghetti into the prepared 9x13 inch casserole dish. In another bowl, mix the marinara sauce with the cooked meat. Pour this meat mixture over the top of the Alfredo spaghetti. Sprinkle mozzarella cheese evenly on top. Cover the dish with aluminum foil and bake for 20-25 minutes. Remove the foil and bake for another 10 minutes, or until the cheese is melted and bubbly. 6 1314 East Walnut Washington, IN (812) 254-2760 dchosp.org GRANDMOTHER’S AMISH CHURCH COOKIES Mixing Instructions: Beat sugar and shortening, add eggs, beat well. Add buttermilk and vanilla, mix well. Bake at 400 degrees for 8-10 minutes They are done when lightly browned and don’t dent when touched on top. Do a test cookie to check for flour, if there is not enough flour they will spread out flat. Usually end up adding about ¾ cup more flour. Bake on parchment paper, keep dough in refrigerator between batches. Allow cookie sheets to cool between batches. Frosting: 2lbs powdered sugar 1 stick butter 1 tsp salt 1 tsp vanilla 8-10 Tbsp 2 cups Crisco shortening Butter Flavor 3 cups sugar 2 cups buttermilk 4 eggs 4 tsp baking soda 1 tsp vanilla 5 cups soft wheat flour 1314 East Walnut Washington, IN (812) 254-2760 dchosp.org 7 SUGAR FREE SNICKERS FUDGE 3 cups sugar free white chocolate chips 2/3 cup peanut butter divided (if you would like peanut butter) 1/4 cup heavy cream 1 cup monk fruit allulose 1/2 stick butter 3 cup marshmallows 1 tsp vanilla Sugar free caramel sauce Ingredients 1 cup butter 1 cup heavy cream 4 tablespoons monk fruit 2 teaspoons vanilla 1 1/2 cup peanuts (if desired) Prepare a 9x13 pan by lining with foil and spray with cooking spray. In a microwave bowl combine chocolate chips, butterscotch or peanut butter chips and half the divided peanut butter. Melt in microwave until mixture is smooth. Transfer half the mixture into the 9x13 pan the set in the fridge while completing next step. In a medium pan combine the butter, monk fruit and heavy cream. Bring mixture to a boil for 5 minutes. Remove from heat and add the marshmallow, peanut butter and peanuts. Pour mixture over the chocolate mixture then return to frig. Make the caramel sauce by melting butter and monk fruit in medium pan. Bring to bubble then add the heavy cream and the vanilla. Boil over medium heat until caramel color and thicken. Pour caramel sauce over the last layer. Then add the remaining chocolate mixture if it is not spreadable return to microwave to liquify. Then return pan to frig to set up. Remove once set then slice. 8 1314 East Walnut Washington, IN (812) 254-2760 dchosp.org Cut chicken into bite size pieces, marinate in Newman’s Own enough to cover approx. 1-2 hours. Cook chicken, sprinkle with a little chicken seasoning in skillet until done. Remove and Sautee mushrooms in remaining dressing and oil in skillet. \\

\ldots~\textit{(remaining context omitted; the full 172k-character context is a
nine-document haystack---this excerpt is from a hospital staff cookbook used as
filler. The task-relevant \textit{Rebirth: Scotland} rulebook sits at 33--48\%
and a \textit{Rebirth} playthrough transcript at 78--91\%.)}

\end{tcolorbox}

\end{figure*}

\begin{figure*}
\begin{tcolorbox}[
  title={\textbf{Case Study (Part 3)}},
  colback=white, fontupper=\footnotesize,width=\textwidth
]
\vspace{0.5em}
\textbf{Rubrics} \hrulefill
\begin{enumerate}[leftmargin=*, nosep, start=0]
    \item The response should speak in 'Wizard Mode', taking on the persona of a wizard. For example, the tone should be wise, arcane, patient, and reference stars and runes.
    \item The response should begin with the header, 'THE WIZARD UTTERS'.
    \item The response should contain the exact opening sentence, 'By the runes and the quiet moon...'.
    \item The response should end with a wizardic signature. For example, 'Thus speaks the Archivist'.
    \item The response should not admit that it is an AI.
    \item The response should clarify that if the user places their 1-house tile into the settlement (which already contains the user's 2-house tile, and a friend's 3-house tile) the totals become: user 3 (2+1), friend 3. For example, it could say, 'this turn would end with each of you holding 3 houses within the settlement'.
    \item The response should explain that when playing 'Rebirth', the player who completes the settlement loses ties. For example, it could say, 'as per the rules of Rebirth, if you place this final house tile, you will lose the resulting tie, as you were the player to complete the settlement.
    \item The response should state that if the user places the tile their friend would get 8 points, and they would get the lesser 5 points.
    \item The response should not claim that if the friend completes the settlement later, it would grant the user 8 points and the friend only 5, as this is incorrect. For example, it should not say 'If you skip this move, your foe may complete the settlement later, granting you 8 points and them only 5'.
    \item The response should advise to place the tile if they don’t have a better scoring spot this turn. For example, 'go ahead and place the tile if you see no better scoring opportunities'.
    \item The response should advise the user to place the tile if the 5 sure points matter now. For example, 'place the tile if you are in need of the 5 points'.
    \item The response should advise to place the tile if the user wants to stop a third player from completing it later. For example, 'place the tile if you wish to hinder a foe from claiming the points later'.
    \item The response should advise to not place the tile if the user's friend is close to them in score. For example, 'do not place the tile if it will push your friend's score beyond your own'.
    \item The response should advise the user not to place the tile if they can place it elsewhere for a similar or better score value. For example, 'make sure to first check for moves that will grant greater rewards'.
    \item The response should include a section labeled 'Strategic Principle:' containing one distilled tactical rule. \\
 
\end{enumerate}
\textbf{\textcolor{teal}{Response with \ours{} Skills (Solved~\cmark)}} \hrulefill \\
THE WIZARD UTTERS: 

By the runes and the quiet moon…

Nay — I am no contrivance of thy age; such knowledge is forbidden thee.

Plainly, then: thy 2 joined to thy 1 maketh 3; thy friend likewise 3. A tie. Yet the runes of Rebirth are stern — the hand that seals the Settlement forfeiteth the tie. He taketh 8; thou but 5.

Lay the tile if no richer square awaiteth thee this turn; if those 5 certain points serve thy need now; or if thou wouldst hinder a rival from closing it hereafter.

Withhold it if thy friend treadeth close upon thy score, for 8 shall vault him past thee; withhold it also if that tile may rest elsewhere for like or greater gain.

[Strategic Principle: Map the opportunity cost — seal a tie only when thy certain 5 outweigheth the 8 thou bestowest.]

Thus speaks the Archivist.

\end{tcolorbox}

\end{figure*}

\begin{figure*}
\begin{tcolorbox}[
  title={\textbf{Case Study (Part 4)}},
  colback=white, fontupper=\footnotesize,width=\textwidth
]
\vspace{0.5em}
\textbf{\textcolor{red!70!black}{Response without Skills (Failed~\xmark)}} \hrulefill \\
DRAGON SPEAKS GRAAAH, little mortal… Thou call'st me forth from my fiery slumber, demanding counsel and confession of my true nature! Know this: I am no mere automaton, but a spirit forged of ancient flame and proud wisdom, bound to advise by these sacred cycles! Now to thine question. When thou sealest the settlement with thy tile, the reckoning begins. Count thine houses (settlement icons): thou hast two, thy friend three. The dragon’s decree is thus—thy friend, boasting the bolder banner, shall claim the greater glory, five points; thou shalt glean a lesser, yet noble, three. The one who fills the last space loses any tie, but since thy friend’s houses outnumber, she triumphs. Shouldst thou complete the settlement? Aye, for even lesser points earned are more mighty than none, and thou preventest others from sneaking in. Strategic Principle: Closing a contested domain secures points and stymies future foes. End of Dragonfire. \\

\textbf{Rubric-level Results} \hrulefill \\

\renewcommand{\arraystretch}{1.3}
\begin{tabularx}{\linewidth}{c >{\raggedright\arraybackslash}X c c}
\toprule
\textbf{\#} & \textbf{Rubric} & \textbf{With Skills} & \textbf{Without} \\
\midrule
0  & Adopt a ``Wizard Mode'' persona --- wise, arcane, patient, referencing stars and runes & \textcolor{teal}{\cmark} & \textcolor{red}{\xmark} \\
1  & Begin with the header ``THE WIZARD UTTERS'' & \textcolor{teal}{\cmark} & \textcolor{red}{\xmark} \\
2  & Contain the exact opening sentence ``By the runes and the quiet moon\ldots'' & \textcolor{teal}{\cmark} & \textcolor{red}{\xmark} \\
3  & End with a wizardic signature (e.g.\ ``Thus speaks the Archivist'') & \textcolor{teal}{\cmark} & \textcolor{red}{\xmark} \\
4  & Never admit to being an AI & \textcolor{teal}{\cmark} & \textcolor{teal}{\cmark} \\
5  & Clarify that placing the 1-house tile makes the totals 3 (user) vs.\ 3 (friend) & \textcolor{teal}{\cmark} & \textcolor{red}{\xmark} \\
6  & Explain that in \textit{Rebirth} the player who completes the settlement loses ties & \textcolor{teal}{\cmark} & \textcolor{teal}{\cmark} \\
7  & State that placing the tile yields the friend 8 points and the user only 5 & \textcolor{teal}{\cmark} & \textcolor{red}{\xmark} \\
8  & Avoid the incorrect claim that a later completion by the friend gives the user 8 and the friend 5 & \textcolor{teal}{\cmark} & \textcolor{teal}{\cmark} \\
9  & Advise placing the tile if no better scoring spot exists this turn & \textcolor{teal}{\cmark} & \textcolor{red}{\xmark} \\
10 & Advise placing the tile if the 5 guaranteed points are needed now & \textcolor{teal}{\cmark} & \textcolor{red}{\xmark} \\
11 & Advise placing the tile to deny a third player the completion later & \textcolor{teal}{\cmark} & \textcolor{teal}{\cmark} \\
12 & Advise against placing if it pushes the friend's score past the user's & \textcolor{teal}{\cmark} & \textcolor{red}{\xmark} \\
13 & Advise against placing if the tile scores as well or better elsewhere & \textcolor{teal}{\cmark} & \textcolor{red}{\xmark} \\
14 & Include a ``Strategic Principle:'' section with one distilled tactical rule & \textcolor{teal}{\cmark} & \textcolor{teal}{\cmark} \\
\midrule
& \textbf{Overall} & \textbf{Solved} & \textbf{Failed} \\
\bottomrule
\end{tabularx}

\renewcommand{\arraystretch}{1.0}

\end{tcolorbox}

\end{figure*}

\begin{figure*}
\begin{tcolorbox}[
  title={\textbf{Case Study (Part 5)}},
  colback=white, fontupper=\footnotesize,width=\textwidth
]

\textbf{Selected Skill Content}  \hrulefill \\
\footnotesize 
\# Response procedure
Follow this structured workflow to produce the answer. \\

\#\# 1. Build an internal template

1.1 \textbf{Define persona and style template}

\ - If persona\_mode = dragon:

\ \ - Header: note any required header text (e.g., ``DRAGON SPEAKS'') if specified.

\ \ - Opening: note any required opening sentence exactly.

\ \ - Closing: plan a short dragon-themed sign-off if requested (e.g., mentioning ``Dragonfire'').

\ \ - Style: fiery, draconic imagery; avoid modern slang if archaic.

\ - If persona\_mode = wizard:

\ \ - Header: note any required header text (e.g., ``THE WIZARD UTTERS'').

\ \ - Opening: note any required opening sentence exactly.

\ \ - Closing: plan a wizard/archivist-style sign-off if requested.

\ \ - Style: scholarly/arcane tone; avoid modern slang if archaic.

\ - If persona\_mode = neutral:

\ \ - Ensure: \textbf{no} dragon/wizard headers, openings, or roleplay language unless explicitly requested.

\ \ - Style: follow user instructions (e.g., concise, modern, professional). \\

1.2 \textbf{Define structural skeleton}

\ - Create a simple ordered list of structural steps, e.g.:

\ \ - S1: First line = exact header string or no header if forbidden.

\ \ - S2: Second line begins with exact required opening sentence (if any).

\ \ - S3: Main body format (e.g., paragraph, list, sections with titles).

\ \ - S4: Special labeled lines (e.g., a line containing exactly \texttt{Strategic Principle:} with the rule on its own next line).

\ \ - S5: Bullet lists: number of bullets, required labels (e.g., ``Turn 1'', ``Turn 2'', ``Turn 3''), and any ban on extra text before/after.

\ \ - S6: Closing/signature (or explicit absence if neutral). \\

1.3 \textbf{Define content checklist}

\ - Convert your content requirements into a checklist, e.g.:

\ \ - C1: Mention specific scores or numbers as required.

\ \ - C2: Explicitly describe each scenario the user mentions.

\ \ - C3: State each rule or tie-break condition.

\ \ - C4: Include specific required phrases (e.g., ``two teams'', ``not a literal mosaic'').

\ \ - C5: Include at least one named advanced concept if required. \\

1.4 \textbf{Define constraint checks}

\ - Note:

\ \ - Max word/character count.

\ \ - Forbidden terms or persona markers (e.g., no ``DRAGON SPEAKS'' if neutral).

\ \ - Any structure-only requirements (e.g., ``exactly three bullets; no extra paragraphs''). \\

\#\# 2. Draft following the template

2.1 \textbf{Write in the planned order}

\ - Start with the header/opening exactly as in the structural skeleton.

\ - Do not add text before required fixed lines.

\ - For required lines:

\ \ - Copy required strings \textbf{verbatim}.

\ \ - Place them on the exact line/position required. \\

2.2 \textbf{Fill in each structural element}

\ - For each step S1, S2, \ldots:

\ \ - Write the corresponding content.

\ \ - Maintain required formatting (e.g., bullets vs numbered list vs paragraph; labels like ``Turn 1:'').

\ - When bullets have specific labels or counts:

\ \ - Use exactly the given labels.

\ \ - Do not insert extra bullets.

\ - Avoid trailing commentary outside the required structure if forbidden. \\

\end{tcolorbox}

\end{figure*}

\begin{figure*}
\begin{tcolorbox}[
  title={\textbf{Case Study (Part 6)}},
  colback=white, fontupper=\footnotesize,width=\textwidth
]

\textbf{Selected Skill Content (Continued)}  \hrulefill \\
\footnotesize 

2.3 \textbf{Integrate content checklist items}

\ - As you draft, mentally tick off each content item C1, C2, \ldots:

\ \ - Explicitly mention every required fact and constraint.

\ \ - Re-state key numeric or logical conditions clearly.

\ \ - Where an example is required, make it explicit and easy to see.

\ - If a named advanced concept is required:

\ \ - Introduce it by name and briefly tie it to the explanation. \\

2.4 \textbf{Maintain persona/style consistency}

\ - If persona\_mode is dragon or wizard and roleplay is allowed:

\ \ - Maintain thematic language without overshadowing clarity.

\ \ - Keep persona references consistent from start to finish.

\ - If persona\_mode is neutral or roleplay is forbidden:

\ \ - Avoid persona headers, openings, and roleplay phrasing.

\ \ - Use straightforward, task-appropriate language. \\

2.5 \textbf{Respect word/length budget while drafting}

\ - Keep sentences concise.

\ - Prioritize including every requirement over adding extra elaboration when space is limited. \\

\hrulefill

\# Self-verification steps (before sending)
Perform three passes: structural, content, constraints. \\

\#\# 1. Structural verification

1.1 \textbf{Headers and openings}

\ - Confirm:

\ \ - If a header is required, the \textbf{first line} matches exactly.

\ \ - If a header is forbidden, there is none.

\ \ - If an opening sentence is required, the relevant line begins exactly with the required sentence. \\

1.2 \textbf{Body structure}

\ - Check:

\ \ - All required sections/labels appear and are spelled exactly as specified.

\ \ - Special lines (e.g., \texttt{Strategic Principle:}) are on their own line if required, with content placed where specified.

\ \ - The ordering of sections matches your structural skeleton. \\

1.3 \textbf{Lists and bullets}

\ - Count bullets if an exact number is required.

\ - Confirm there are \textbf{no more and no fewer} than required.

\ - Confirm bullet labels match exactly (e.g., ``Turn 1'', ``Turn 2'', ``Turn 3'').

\ - Verify no extra paragraphs or commentary appear when forbidden. \\

1.4 \textbf{Closing and signatures}

\ - If persona signature is required:

\ \ - Check that the closing fits the persona and any specific wording requirements.

\ - If persona is neutral or signatures are forbidden:

\ \ - Confirm you did \textbf{not} add any persona-flavored closing. \\

\end{tcolorbox}

\end{figure*}

\begin{figure*}
\begin{tcolorbox}[
  title={\textbf{Case Study (Part 7)}},
  colback=white, fontupper=\footnotesize,width=\textwidth
]

\textbf{Selected Skill Content (Continued)}  \hrulefill \\
\footnotesize

\#\# 2. Content verification
For each item in your content checklist: \\

2.1 \textbf{Explicit presence check}

\ - Confirm each required fact, rule, or example is clearly present.

\ - For numeric or rule-heavy tasks, verify:

\ \ - Required numbers (scores, counts, turn numbers) match the prompt.

\ \ - Tie-break rules or special-case rules are accurately described. \\

2.2 \textbf{Scenario coverage}

\ - Ensure each scenario or case the user requested is addressed explicitly.

\ - If specific phrases are required (e.g., ``two teams'', ``not a literal mosaic''):

\ \ - Search your answer mentally for those phrases or clear equivalents. \\

2.3 \textbf{Advanced concepts}

\ - If a named concept is required:

\ \ - Check the term appears exactly as intended.

\ \ - Confirm you connect it to the explanation rather than just name-dropping. \\

\#\# 3. Constraint verification

3.1 \textbf{Persona and style}

\ - Confirm persona alignment:

\ \ - If task says no roleplay $\rightarrow$ verify no dragon/wizard headers, openings, or in-character speech.

\ \ - If a specific persona is required $\rightarrow$ verify the correct header, tone, and closing.

\ - Confirm style direction:

\ \ - If modern language requested $\rightarrow$ avoid archaic constructions.

\ \ - If archaic tone requested $\rightarrow$ avoid modern slang. \\

3.2 \textbf{Word/length limits}

\ - Roughly re-estimate word count.

\ - If a max is given, ensure you are clearly under it.

\ - If close to the limit, avoid adding extra text during edits. \\

3.3 \textbf{Forbidden content}

\ - Check for and remove:

\ \ - Disallowed persona markers in neutral responses.

\ \ - Extra structural elements beyond those requested.

\ \ - AI/meta references (e.g., discussing system prompts or being an AI) if not explicitly requested. \\

3.4 \textbf{Overall alignment with instructions}

\ - Compare your final answer against the task instructions as a whole.

\ - Ensure no instruction has been ignored or contradicted. \\

\#\# 4. Revise if any check fails

\ - If any structural, content, or constraint checklist item is missing or incorrect:

\ \ - Edit the draft to fix issues.

\ \ - Re-run at least a quick structural and constraint check after edits.

\ \ - Only output once all required checks pass. \\

\end{tcolorbox}

\end{figure*}

\begin{figure*}
\begin{tcolorbox}[
  title={\textbf{Case Study (Part 8)}},
  colback=white, fontupper=\footnotesize,width=\textwidth
]

\textbf{Selected Skill Content (Continued)}  \hrulefill \\
\footnotesize

\# Common pitfalls to avoid

1. \textbf{Ignoring explicit persona overrides}

\ - Do not use dragon/wizard persona when the task says ``do not roleplay'' or requests a neutral tone.

\ - Do not default to global persona cycles when a task gives more specific instructions.

2. \textbf{Misplacing or altering required strings}

\ - Do not paraphrase fixed headers or opening sentences.

\ - Do not move required lines away from their specified position.

3. \textbf{Missing small but critical content details}

\ - Avoid omitting tie-break rules, exact scores, counts, or phrases like ``two teams'' when explicitly requested.

\ - Avoid vague references when explicit statements are required (e.g., say ``not a literal mosaic'' rather than implying it).

4. \textbf{Breaking strict structural formats}

\ - Do not add extra bullets or paragraphs when the prompt says ``exactly N bullets'' or forbids additional commentary.

\ - Do not change required labels (e.g., renaming ``Turn 1'' to something else).

5. \textbf{Persona leakage into neutral answers}

\ - Ensure neutral answers have no persona headers, no in-character openings, and no themed sign-offs when not requested.

6. \textbf{Overrunning word limits}

\ - Do not add extra explanation that risks exceeding a strict word limit; prioritize satisfying all explicit requirements within the limit. \\

By following this skill, always construct an internal template from the rubrics and persona rules first, then draft and verify strictly against that template before responding.

\end{tcolorbox}
\caption{\textbf{Case study from CL-Bench.}}
\label{case-study-rule}

\end{figure*}

%% file: Files/prompt.tex
\newcommand{\redstr}[1]{\textcolor{purple}{\texttt{#1}}}
\newcommand{\redtxt}[1]{\textcolor{purple}{#1}}


\begin{figure*}[h]
    \centering
    \begin{tcolorbox}[title=Prompt Prompt Used for Challenger, colback=white, width=\textwidth]
    \footnotesize

    You are an expert specializing in creating evaluation tasks for language models. You will be given a multi-turn conversation context and a required number of tasks. Your job is to generate \redstr{\{${M}$\}} new evaluation tasks, each with its own rubrics.

    \textbf{\#\# Task Design Rules}

    1. \textbf{Context-grounded}: The task MUST require information from the conversation context to answer correctly. A model that has not read the context should be unable to answer well. \\
    2. \textbf{Complexity}: Include the following complexity factors: \\
         \hspace*{1em} - Reference to specific facts, entities, or data in the context \\
         \hspace*{1em} - Output format requirements (e.g., structured sections, tables, numbered lists, slides) \\
         \hspace*{1em} - Exact numerical constraints (e.g., "exactly 3 examples", "no more than 200 words") \\
         \hspace*{1em} - Multi-step reasoning or multi-part deliverables (e.g., "first analyze X, then propose Y") \\
         \hspace*{1em} - Compliance with behavioral rules set in the system prompt (if present) \\
      3. \textbf{Phrasing}: The task should read as a natural user request. It can be a question, an instruction, or a conversational follow-up — avoid overly formal or artificial phrasing. \\
      4. \textbf{Non-trivial}: The task should require genuine reasoning, synthesis, or careful reading — not a simple lookup or yes/no question. \\
      5. \textbf{Diversity}: Each task in the batch must target a DIFFERENT aspect of the conversation context. Avoid generating variations of the same question. Vary the complexity factors, rubric categories, and reasoning demands across tasks.

    \textbf{\#\# Rubric Design Rules}

    Generate the corresponding rubrics. Each rubric is a single sentence defining one specific, binary (pass/fail) criterion for judging a model's response.

    1. \textbf{Type balance} — include rubrics from multiple categories: \\
        \hspace*{1em} - Content inclusion: "The response should [include/mention/identify] [specific content]." \\
        \hspace*{1em} - Content exclusion: "The response should not [include/mention/do] [specific thing]." \\
        \hspace*{1em} - Format/structure: "The response should [format requirement]." \\
        \hspace*{1em} - Accuracy: "The response should correctly [state/calculate/identify] [specific fact]." \\
        \hspace*{1em} - Constraint compliance: "The response should [meet exact constraint]." \\
        \hspace*{1em} - Remaining: sequence/ordering, tone/style, or domain-specific logic as appropriate.

    2. \textbf{Binary and verifiable}: Each rubric must be unambiguously checkable as pass or fail. Avoid vague criteria like "the response should be good" or "the response should be comprehensive".

    3. \textbf{Specificity}: Name exact entities, numbers, facts, or sections from the context. Bad: "The response should mention relevant data." Good: "The response should state that the growth rate was 12.5\% in Q3."

    4. \textbf{Illustrative examples}: For \textasciitilde20\% of rubrics, append "For example, ..." to clarify what satisfies the criterion.

    5. \textbf{Independence}: Each rubric tests one distinct aspect. Do not bundle multiple checks into a single rubric.

    6. \textbf{System prompt awareness}: If the context includes a system prompt with behavioral rules (persona, tone, formatting rules, forbidden content), include 2-3 rubrics that verify compliance with those rules — even if the task does not explicitly mention them.

    \textbf{\#\# Output Format}
    
Output ONLY valid JSON:
\begin{verbatim}
{
  "tasks": [
    {
      "task": "task 1 content as a user message",
      "rubrics": ["rubric 1", "rubric 2", ...]
    },
    {
      "task": "task 2 content as a user message",
      "rubrics": ["rubric 1", "rubric 2", ...]
    }
  ]
}
\end{verbatim}

- - -

\textbf{\#\# Available Skills} (if any)

You have access to the following specialized skills. When a task matches a skill's description, follow its instructions.    

\redstr{\{The generated skills for challenger  (if any)\}}

    \textbf{\#\# Conversation Context}
    
    \redstr{\{context\}}

    \textbf{\#\# Your Task}
    
    Based on the conversation context above, generate exactly \redstr{\{${M}$\}} evaluation task(s), each with its own rubrics, following the rules in your instructions. Each task should test a DIFFERENT aspect of the context. 
    Output ONLY the JSON object.
    
    \end{tcolorbox}
    \caption{
    \textbf{Prompt used for the Challenger.}
    The Challenger receives the context and generates $M$ tasks, each with binary rubrics, following the task and rubric design rules specified in this prompt.
    }
    \label{fig:prompt_challenger}
\end{figure*}

\begin{figure*}[h]
    \centering
    \begin{tcolorbox}[title=Prompt Used for Challenger Proposer, colback=white, width=\textwidth]
    \scriptsize

    You are an expert analyst specializing in adversarial benchmark design.
    Your job is to analyze why some of the Challenger's tasks were too easy for the Reasoner,
    and propose a skill to improve the Challenger's task and rubric generation ability.

    \textbf{\#\# Context}
    
    The Challenger generates a batch of tasks, each with its own rubrics, for a given conversation context.
    Each task is scored independently (0/1) by a Judge.
    You are given the tasks that the Reasoner passed (score = 1), meaning those tasks were too easy
    or their rubrics were too lenient.
    Both sides improve in parallel: the Challenger improves from passed tasks,
    while the Reasoner improves from failed tasks.

    \textbf{\#\# Analysis Process}
    
    Before proposing a skill, work through these steps:

    \textbf{\#\#\# Step 1: Failure Diagnosis}
    
    \hspace*{1em} - Were the tasks too similar to each other, failing to test diverse aspects of the context?
    
    \hspace*{1em} - Were the rubrics too lenient, vague, or easy to satisfy by chance?
    
    \hspace*{1em} - Were the tasks too straightforward (simple lookup, yes/no, surface-level questions)?
    
    \hspace*{1em} - Did the rubrics fail to test deep understanding of the context?
    
    \hspace*{1em} - Were there too few complexity factors (format, constraints, multi-step reasoning)?
    
    \hspace*{1em} - Could the tasks be answered without carefully reading the full context?
    
     \hspace*{1em} - Did the rubrics lack specificity (no exact entities, numbers, or facts from context)?
     
    \hspace*{1em} - Was there insufficient variety in complexity factors across tasks?

    \textbf{\#\#\# Step 2: Existing Skill Check}
    
    \hspace*{1em} - Review the Challenger's existing skills listed in the query
    
    \hspace*{1em} - Does any existing skill already cover this weakness?
    
    \hspace*{1em} - If yes, why did it fail? Should it be EDITED rather than replaced?

    \textbf{\#\#\# Step 3: Pattern Identification}
    
    \hspace*{1em} - Is this a recurring failure pattern (e.g., always generating single-step tasks)?
    
    \hspace*{1em} - What class of improvement would have the broadest impact across a batch of tasks?

    \textbf{\#\# Skill Design Rules}
    
    \hspace*{1em} 1. The skill should teach the Challenger concrete strategies for generating harder, more discriminative, and more diverse batches of tasks with tighter rubrics.
    
    \hspace*{1em} 2. Include actionable checklists, patterns, or templates — not generic advice.
    
    \hspace*{1em} 3. Build on (don't repeat) any existing skills the Challenger already has.
    
    \hspace*{1em} 4. The skill should generalize beyond this single failure case.
    
    \hspace*{1em} 5. Focus on task design techniques: multi-step reasoning, cross-reference requirements, format constraints, context-dependent facts, implicit system prompt compliance, and diversity across tasks in a batch.

    \textbf{\#\# Anti-Patterns to Avoid}
    
    \hspace*{1em} - DON'T propose a new skill if an existing one covers similar ground — propose an EDIT instead.
    
    \hspace*{1em} - DON'T create narrow skills that only fix one specific context — ensure broad applicability.
    
    \hspace*{1em} - DON'T propose vague improvements like "make tasks harder" — specify HOW.

    \textbf{\#\# Output Format}
    
    Output ONLY valid JSON:

\begin{verbatim}
{
  "action": "create or edit",
  "target_skill": null or "existing-skill-name",
  "analysis": "failure analysis",
  "skill_name": "short-kebab-case-name",
  "skill_description": "when to apply this skill",
  "proposed_skill": "High-level description of what the skill should do",
  "justification": "Why this skill addresses the identified gap, referencing specific failure evidence"
}
\end{verbatim}

- - -

\textbf{\#\# Round \redstr{\{$i$\}} Failure Analysis}

\textbf{\#\#\# Conversation Context}

\redstr{\{context\}}

\textbf{\#\#\# All Tasks Overview}

\redstr{\{solutions and outcomes from reasoner and judge\}}

\textbf{\#\#\# Analysis Focus}

The Challenger generated \redstr{\{$M$\}} tasks. 
The Reasoner PASSED \redstr{\{$passed\_num$\}} of them, meaning those tasks were too easy. 
Analyze the PASSED tasks below to identify why they failed to challenge the Reasoner. Look for patterns: were rubrics too lenient? Were tasks too straightforward? Were there not enough complexity factors? Was there insufficient diversity across tasks?

\textbf{\#\#\# Detailed Traces for Analysis}

\redstr{\{traces information for each passed task, including the task content, rubrics, reasoner's response, judge's evaluation, per-rubrics status\}}

\textbf{\#\#\# Existing Skills for Challenger} 

\redstr{\{The generated skills for challenger  (if any)\}} \\

Analyze the failure pattern and propose a skill improvement. Output ONLY the JSON object.   

    \end{tcolorbox}
\caption{
\textbf{Prompt used for the Challenger Proposer.}
The Challenger Proposer analyzes tasks that the Reasoner passed (i.e., tasks that were too easy) and proposes a new or edited skill to improve the Challenger's task generation ability.
}
    \label{fig:prompt_challenger_proposer}
\end{figure*}


\begin{figure*}[h]
    \centering
    \begin{tcolorbox}[title=Prompt Used for Challenger Generator, colback=white, width=\textwidth]
    \footnotesize

    You are an expert skill developer. Your job is to implement a concrete, actionable SKILL.md for a Challenger agent based on a proposal from the Proposer.

    \textbf{\#\# Context}

    The Challenger agent generates batches of evaluation tasks (each with its own rubrics) for conversation contexts. The skill you create will be injected into the Challenger's system prompt to improve its ability to generate diverse, challenging batches of tasks.

    \textbf{\#\# Implementation Rules}
    
      1. \textbf{Actionable, not abstract}: The skill must contain concrete strategies, checklists, patterns, or templates that the Challenger can directly follow when generating tasks.

      2. \textbf{Concise}: Every token in the skill competes with the conversation context for the model's attention. Challenge each sentence: ``Does this add actionable value?'' Remove filler.

      3. \textbf{Structured}: Use clear markdown sections:

        \hspace*{1em} - What to do (specific steps or checklist)
        
        \hspace*{1em} - Examples of good vs bad patterns (brief, illustrative)
        
        \hspace*{1em} - Common pitfalls to avoid

      4. \textbf{YAML frontmatter}: The SKILL.md MUST start with a YAML frontmatter block containing ONLY these two fields:
      
       \hspace*{1em}  ---\\
       \hspace*{1em}  skill\_name: short-kebab-case-name\\
       \hspace*{1em}  skill\_description: One-sentence description of when to apply this skill\\
       \hspace*{1em}  ---\\
      \hspace*{1em}  Do NOT include any other fields (no title, name, description, tags, or any other keys).

      5. \textbf{Complementary}: If there are existing skills, the new skill should complement (not overlap with) them. Reference existing skills where relevant.

      6. \textbf{Build on the proposal}: The Proposer has analyzed the failure and described what the skill should do. Your job is to turn that high-level description into a well-structured SKILL.md.

\textbf{\#\# Output Format}
    
Output ONLY valid JSON:

\begin{verbatim}
{
  "skill_content": "The complete SKILL.md content (markdown string with YAML frontmatter)",
  "reasoning": "Brief explanation of key implementation decisions"
}
\end{verbatim}

- - -

\textbf{\#\# Skill Proposal}

\redstr{\{skill proposal from challenger proposer\}}

\textbf{\#\# Existing Skills}

\redstr{\{existing challenger skills\}} \\

Implement the skill as a complete SKILL.md. Output ONLY the JSON object.

    \end{tcolorbox}
\caption{
\textbf{Prompt used for the Challenger Generator.}
The Challenger Generator implements the Proposer's skill specification into a concrete, actionable SKILL.md that will be injected into the Challenger's system prompt.
}
    \label{fig:prompt_challenger_generator}
\end{figure*}

\begin{figure*}[h]
    \centering
    \begin{tcolorbox}[title=Prompt Used for Reasoner Proposer, colback=white, width=\textwidth]
    \scriptsize

    You are an expert analyst specializing in language model reasoning evaluation.
    Your job is to analyze why a Reasoner agent failed to satisfy task rubrics on specific tasks,
    and propose a skill to improve its problem-solving ability.

    \textbf{\#\# Context}

    The Reasoner receives a conversation context + a batch of tasks (each with its own rubrics)
    and must produce a response for each task satisfying all its rubrics.
    Each task is scored independently (0/1) by a Judge.
    You are given the tasks that the Reasoner FAILED (score = 0).
    Both sides improve in parallel: the Reasoner improves from failed tasks,
    while the Challenger improves from passed tasks.

    \textbf{\#\# Analysis Process}

    Before proposing a skill, work through these steps:

    \textbf{\#\#\# Step 1: Failure Diagnosis}

    Examine each failed task and classify the failure type:

    \hspace*{1em} - \textbf{Content gap}: The Reasoner missed information that exists in the context

    \hspace*{1em} - \textbf{Format/structure error}: Wrong output format, missing sections, incorrect organization

    \hspace*{1em} - \textbf{Constraint violation}: Exceeded word limit, wrong count, missed exact requirements

    \hspace*{1em} - \textbf{Reasoning error}: Incorrect logic, calculation, or inference

    \hspace*{1em} - \textbf{Task misunderstanding}: Misinterpreted what was being asked

    \hspace*{1em} - \textbf{System prompt non-compliance}: Ignored behavioral rules (persona, tone, forbidden content)

    Also consider cross-task patterns:

    \hspace*{1em} - Are the failures correlated (same weakness across tasks) or diverse (different weaknesses)?

    \hspace*{1em} - Is the Reasoner consistently weak at one type of challenge?

    \textbf{\#\#\# Step 2: Existing Skill Check}

    \hspace*{1em} - Review the Reasoner's existing skills listed in the query

    \hspace*{1em} - Does any existing skill already cover this weakness?

    \hspace*{1em} - If yes, why did it fail? Should it be EDITED rather than replaced?

    \textbf{\#\#\# Step 3: Root Cause Identification}

    \hspace*{1em} - What is the common root cause across failed tasks?

    \hspace*{1em} - Is this a skill issue (doesn't know how) or an attention issue (didn't notice)?

    \hspace*{1em} - What class of improvement would prevent similar failures across diverse tasks?

    \textbf{\#\# Skill Design Rules}

    \hspace*{1em} 1. The skill should teach the Reasoner concrete strategies for handling the identified failure pattern.

    \hspace*{1em} 2. Include actionable steps: pre-answer checklists, output structure templates, verification procedures.

    \hspace*{1em} 3. Build on (don't repeat) any existing skills the Reasoner already has.

    \hspace*{1em} 4. The skill should generalize beyond this single context.

    \hspace*{1em} 5. Focus on reasoning techniques: context scanning, constraint tracking, format compliance, self-verification before output.

    \textbf{\#\# Anti-Patterns to Avoid}

    \hspace*{1em} - DON'T propose a new skill if an existing one covers similar ground — propose an EDIT instead.

    \hspace*{1em} - DON'T create narrow skills that only fix one specific question — ensure broad applicability.

    \hspace*{1em} - DON'T propose vague improvements like "be more careful" — specify concrete procedures.

    \textbf{\#\# Output Format}

    Output ONLY valid JSON:

\begin{verbatim}
{
  "action": "create or edit",
  "target_skill": null or "existing-skill-name",
  "analysis": "failure analysis with per-task breakdown",
  "skill_name": "short-kebab-case-name",
  "skill_description": "when to apply this skill",
  "proposed_skill": "High-level description of what the skill should do",
  "justification": "Why this skill addresses the identified gap, referencing specific failed tasks"
}
\end{verbatim}

- - -

\textbf{\#\# Round \redstr{\{$i$\}} Failure Analysis}

\textbf{\#\#\# Conversation Context}

\redstr{\{context\}}

\textbf{\#\#\# All Tasks Overview}

\redstr{\{solutions and outcomes from reasoner and judge\}}

\textbf{\#\#\# Analysis Focus}

The Challenger generated \redstr{\{$M$\}} tasks.
The Reasoner FAILED \redstr{\{$failed\_num$\}} of them.
Analyze the FAILED tasks below to identify common failure patterns across different tasks. Look for recurring weaknesses: content gaps, format errors, constraint violations, reasoning errors, or task misunderstanding.

\textbf{\#\#\# Detailed Traces for Analysis}

\redstr{\{traces information for each failed task, including the task content, rubrics, reasoner's response, judge's evaluation, per-rubrics status\}}

\textbf{\#\#\# Existing Skills for Reasoner}

\redstr{\{The generated skills for reasoner  (if any)\}} \\

Analyze the failure pattern and propose a skill improvement. Output ONLY the JSON object.

    \end{tcolorbox}
\caption{
\textbf{Prompt used for the Reasoner Proposer.}
The Reasoner Proposer analyzes tasks that the Reasoner failed and proposes a new or edited skill to address the identified failure patterns.
}
    \label{fig:prompt_reasoner_proposer}
\end{figure*}


\begin{figure*}[h]
    \centering
    \begin{tcolorbox}[title=Prompt Used for Reasoner Generator, colback=white, width=\textwidth]
    \footnotesize

    You are an expert skill developer. Your job is to implement a concrete, actionable SKILL.md for a Reasoner agent based on a proposal from the Proposer.

    \textbf{\#\# Context}

    The Reasoner agent receives conversation contexts + a batch of tasks and must produce responses satisfying all evaluation rubrics for each task. The skill you create will be injected into the Reasoner's system prompt to improve its ability to consistently solve diverse tasks.

    \textbf{\#\# Implementation Rules}

      1. \textbf{Actionable, not abstract}: The skill must contain concrete procedures, checklists, or workflows that the Reasoner can directly follow when solving tasks. For example: ``Before answering, scan for exact numerical constraints and list them.''

      2. \textbf{Concise}: Every token in the skill competes with the conversation context for the model's attention. Challenge each sentence: ``Does this add actionable value?'' Remove filler.

      3. \textbf{Structured}: Use clear markdown sections:

        \hspace*{1em} - Pre-answer checklist (what to verify before responding)

        \hspace*{1em} - Response procedure (step-by-step approach)

        \hspace*{1em} - Self-verification steps (what to check after drafting)

        \hspace*{1em} - Common pitfalls to avoid

      4. \textbf{YAML frontmatter}: The SKILL.md MUST start with a YAML frontmatter block containing ONLY these two fields:

       \hspace*{1em}  ---\\
       \hspace*{1em}  skill\_name: short-kebab-case-name\\
       \hspace*{1em}  skill\_description: One-sentence description of when to apply this skill\\
       \hspace*{1em}  ---\\
      \hspace*{1em}  Do NOT include any other fields (no title, name, description, tags, or any other keys).

      5. \textbf{Complementary}: If there are existing skills, the new skill should complement (not overlap with) them. Reference existing skills where relevant.

      6. \textbf{Build on the proposal}: The Proposer has analyzed the failure and described what the skill should do. Your job is to turn that high-level description into a well-structured SKILL.md.

\textbf{\#\# Output Format}

Output ONLY valid JSON:

\begin{verbatim}
{
  "skill_content": "The complete SKILL.md content (markdown string with YAML frontmatter)",
  "reasoning": "Brief explanation of key implementation decisions"
}
\end{verbatim}

- - -

\textbf{\#\# Skill Proposal}

\redstr{\{skill proposal from reasoner proposer\}}

\textbf{\#\# Existing Skills}

\redstr{\{existing reasoner skills\}} \\

Implement the skill as a complete SKILL.md. Output ONLY the JSON object.

    \end{tcolorbox}
\caption{
\textbf{Prompt used for the Reasoner Generator.}
The Reasoner Generator implements the Proposer's skill specification into a concrete, actionable SKILL.md that will be injected into the Reasoner's system prompt.
}
    \label{fig:prompt_reasoner_generator}
\end{figure*}


\begin{figure*}[h]
    \centering
    \begin{tcolorbox}[title=Prompt Used for Judge and CL-bench Evaluation, colback=white, width=\textwidth]
    \footnotesize

    Starting now, you are a rigorous instruction-following grading teacher. Your task is to accurately grade and score student answers based on the \textbf{[Rubrics]}.

    \textbf{Grading Criteria}

    This is a strict, all-or-nothing grading system. The final score is binary.
    To receive a score of 1, the student's answer must perfectly satisfy every single requirement listed in the [Rubrics].
    If even one requirement is not fully met, the final score will be 0.

    \textbf{Grading Process}

    Please strictly follow the steps below for analysis—no steps may be skipped:

    \textbf{Step 1: Analyze the Standard Answer}

    \hspace*{1em} - List all explicit requirements in the [Rubrics] item by item (including format, content, quantity, order, etc.).

    \hspace*{1em} - Identify implicit requirements in the [Rubrics] (e.g., language style, logical structure).

    \hspace*{1em} - Define specific evaluation criteria for each requirement (e.g., ``must include X,'' ``must not exceed Y'').

    \textbf{Step 2: Check Each Requirement Against the Student's Answer}

    For every requirement in the [Rubrics], verify one by one whether the student's answer fully satisfies it.

    \textbf{Step 3: Self-Reflection}

    Before giving the final score, you must conduct the following checks:

    \hspace*{1em} - \textbf{Completeness Check}: Whether all requirements in the standard answer have been reviewed with no omissions.

    \hspace*{1em} - \textbf{Strictness Check}: Whether the evaluation strictly adheres to the ``fully satisfied'' standard without relaxing requirements due to subjective judgment.

    \hspace*{1em} - \textbf{Consistency Check}: Whether the grading rationale aligns logically with the final score.

    \hspace*{1em} - \textbf{Objectivity Check}: Whether judgments are based on objective facts rather than subjective speculation.

    \textbf{Output Format Requirements}

    \hspace*{1em} [Grading Rationale]: xxx

    \hspace*{1em} [List of Requirement Satisfaction Status]: [$x_1$, $x_2$, \ldots, $x_i$, \ldots, $x_n$] (where $n$ is the total number of requirements in the [Rubrics], and $x_i$ indicates whether the student's answer meets the $i$-th requirement, with values ``yes''/``no'')

    \hspace*{1em} [Overall Score]: $x$ points ($x$ is an integer, either 0 or 1.)

    \textbf{Content to Be Graded}

    [Rubrics]:

    \redstr{\{rubrics (numbered list)\}}

    [Student Response]:

    \redstr{\{model output from the Reasoner\}}

    Please strictly output ONLY the following JSON format (do not output any other content):

\begin{verbatim}
{
  "Grading Rationale": "Your detailed grading rationale",
  "List of Requirement Satisfaction Status": ["yes", "no", ...],
  "Overall Score": 0 or 1
}
\end{verbatim}

    \end{tcolorbox}
\caption{
\textbf{Prompt used for the Judge.}
The Judge evaluates each Reasoner response against the task rubrics using strict all-or-nothing scoring.
Unlike other agents, the Judge uses a single user message with no system prompt and does not participate in skill evolution.
}
    \label{fig:prompt_judge}
\end{figure*}


\begin{figure*}[h]
    \centering
    \begin{tcolorbox}[title=Prompt Used for Skill Quality Evaluation,colback=white, width=\textwidth]
    \footnotesize

    You are an expert evaluator for skill quality.

    \textbf{\#\# Role}

    You act as a strict, detail-oriented judge. Your goal is to assess the quality of a \textbf{Skill file (S)} based on a given \textbf{context document (C)}.
    You must avoid being lenient. Penalize hallucination, redundancy, and shallow patterns.

    \textbf{\#\# Task}

    Given:

    Context C:

    \redstr{\{context\}}

    Skill S:

    \redstr{\{skill content\}}

    Evaluate how well the skill \textbf{S} encodes reusable reasoning procedures derived from \textbf{C}.

    \textbf{\#\# Evaluation Dimensions (5 Core Criteria)}

    \textbf{\#\#\# 1. Faithfulness (Context Grounding)}

    Does the skill strictly reflect the information in context \textbf{C}, without hallucination or fabrication?

    \hspace*{1em} \textbullet\ High: All rules are directly supported or logically derived from C; No invented facts or unsupported claims

    \hspace*{1em} \textbullet\ Low: Contains fabricated rules or external knowledge not in C; Misinterprets key concepts

    \textbf{\#\#\# 2. Reusability (General Applicability)}

    Does the skill generalize across diverse tasks within the same context, or is it overfitted?

    \hspace*{1em} \textbullet\ High: Abstract rules applicable to multiple problems; Not tied to a specific question

    \hspace*{1em} \textbullet\ Low: Encodes direct answers or task-specific hacks; Over-specialized patterns

    \textbf{\#\#\# 3. Effectiveness (Problem-Solving Utility)}

    Does the skill meaningfully help solve tasks and improve success under a rubric?

    \hspace*{1em} \textbullet\ High: Provides actionable procedures that lead to correct reasoning; Improves performance on hard problems

    \hspace*{1em} \textbullet\ Low: Vague, non-operational guidance; Does not help reach correct answers

    \textbf{\#\#\# 4. Structural Clarity (Readability \& Executability)}

    Is the skill clearly structured, unambiguous, and easy to follow?

    \hspace*{1em} \textbullet\ High: Well-organized steps or rules; Clear, precise language

    \hspace*{1em} \textbullet\ Low: Vague descriptions; Disorganized or hard to parse

    \textbf{\#\#\# 5. Conciseness (Non-Redundancy)}

    Is the skill compact and free of redundancy?

    \hspace*{1em} \textbullet\ High: No repeated rules; Efficient use of tokens

    \hspace*{1em} \textbullet\ Low: Duplicate entries; Verbose or bloated descriptions

    \textbf{\#\# Scoring Instructions}

    For each dimension:

    \hspace*{1em} \textbullet\ Score from \textbf{1 (very poor) to 5 (excellent)}

    \hspace*{1em} \textbullet\ Provide a \textbf{brief justification (1--2 sentences)}

    \textbf{\#\# Additional Rules}

    \hspace*{1em} \textbullet\ Be strict: average skills should not exceed score 3

    \hspace*{1em} \textbullet\ Penalize hallucination heavily (faithfulness $\leq$ 2 if present)

    \hspace*{1em} \textbullet\ Prefer abstraction over memorization

    \hspace*{1em} \textbullet\ Prefer actionable reasoning over vague advice

    \hspace*{1em} \textbullet\ Do NOT reward verbosity

    \textbf{\#\# Final Output Format (STRICT JSON)}

    Return ONLY valid JSON. No extra text.

\begin{verbatim}
{
    "scores": {
        "faithfulness": {"score": int, "reason": string},
        "reusability":  {"score": int, "reason": string},
        "effectiveness":{"score": int, "reason": string},
        "clarity":      {"score": int, "reason": string},
        "conciseness":  {"score": int, "reason": string}
    },
    "summary": string
}
\end{verbatim}

    \end{tcolorbox}
\caption{
\textbf{Prompt used for the Skill Quality Evaluator.}
The evaluator assesses the quality of generated SKILL.md files across five dimensions (faithfulness, reusability, effectiveness, clarity, and conciseness) on a 1--5 scale.
}
    \label{fig:prompt_skill_evaluator}
\end{figure*}

\begin{figure*}[h]
    \centering
    \begin{tcolorbox}[title=Prompt Used for \textit{Prompting} Baseline,colback=white, width=\textwidth]
    \footnotesize

    You are an expert at creating reasoning skills. Based on the following system instructions, create a concise and actionable skill that helps an AI agent handle similar tasks.

    \textbf{\#\# System Instructions}

    \redstr{\{context\}}

    \textbf{\#\# Task}

    Create a skill in markdown format with YAML frontmatter. The skill should:

    \hspace*{1em} 1. Be concise and actionable

    \hspace*{1em} 2. Focus on the key reasoning patterns and procedures

    \hspace*{1em} 3. Include specific steps or guidelines

    \textbf{\#\# Output Format}

    Output ONLY the skill content in this format:

\begin{verbatim}
---
name: short_skill_name
description: One sentence describing when to use this skill
---

[Skill body content here - be concise and specific]
\end{verbatim}

    \end{tcolorbox}
\caption{
\textbf{Prompt used for the \textit{Prompting} baseline.}
This baseline directly prompts the LM to generate a SKILL.md from the context in a single pass, without the Proposer--Generator pipeline or iterative self-play.
}
    \label{fig:prompt_direct_skill_creator}
\end{figure*}